\documentclass[paper=letter,DIV=10]{scrartcl}

\usepackage{amsmath,amsfonts,amssymb,amsthm,bm,mathtools,bbm}
\usepackage{subcaption}
\usepackage{graphicx}
\usepackage{tabularray}
\usepackage{multirow}
\usepackage{graphicx}
\usepackage{array}
\usepackage{xcolor}
\usepackage[numbers]{natbib}
\usepackage{url}
\usepackage{authblk}

\DeclarePairedDelimiterX{\infdivx}[2]{(}{)}{%
  #1\;\delimsize\|\;#2%
}
\newcommand{\kldiv}{D_{\text{KL}}\infdivx}

\title{Variational Encoder-Decoders for Learning Latent Representations of Physical Systems}
\author[1]{Subashree Venkatasubramanian}
\author[2]{David A. Barajas-Solano\thanks{Corresponding author: David.Barajas-Solano@pnnl.gov (David A. Barajas-Solano)}}
\affil[1]{Columbia University, New York, NY 10027}
\affil[2]{Pacific Northwest National Laboratory, Richland, WA 99354}
\date{}

\begin{document}

\maketitle

\begin{abstract}
  We present a deep-learning Variational Encoder-Decoder (VED) framework for learning data-driven low-dimensional representations of the relationship between high-dimensional parameters of a physical system and the system’s high-dimensional observable response. The framework consists of two deep learning-based probabilistic transformations: An encoder mapping parameters to latent codes and a decoder mapping latent codes to the observable response. The hyperparameters of these transformations are identified by maximizing a variational lower bound on the log-conditional distribution of the observable response given parameters. To promote the disentanglement of latent codes, we equip this variational loss with a penalty on the off-diagonal entries of the aggregate distribution covariance of codes. This regularization penalty encourages the pushforward of a standard Gaussian distribution of latent codes to approximate the marginal distribution of the observable response.

  Using the proposed framework we successfully model the hydraulic pressure response at observation wells of a groundwater flow model as a function of its discrete log-hydraulic transmissivity field. Compared to the canonical correlation analysis encoding, the VED model achieves a lower-dimensional latent representation, with as low as $r = 50$ latent dimensions without a significant loss of reconstruction accuracy. We explore the impact of regularization on model performance, finding that KL-divergence and covariance regularization improve feature disentanglement in latent space while maintaining reconstruction accuracy. Furthermore, we evaluate the generative capabilities of the regularized model by decoding random Gaussian noise, revealing that tuning both $\beta$ and $\lambda$ parameters enhances the quality of the generated observable response data.

\end{abstract}

\section{Introduction}

The map from parameters of a physical system to the system's observable response often exhibits a ``low-dimensional'' structure due to the spatiotemporal correlations between the degrees of freedom of the system's parameters and response. For such problems, a limited number $r$ of either parameter space coordinates or possibly nonlinear transformations of the parameters capture most of the variation of the observable response, with $r$ much less than the ambient dimensions $n$ and $m$ of the parameter and response spaces, respectively.
More formally, a map $h \colon \mathbb{R}^n \to \mathbb{R}^m$ satisfying this low-dimensional hypothesis can be accurately approximated as the composition of an ``encoding'' transformation $g \colon \mathbb{R}^n \to Z \coloneqq \mathbb{R}^r$ and a ``decoding'' transformation $f \colon Z \to \mathbb{R}^m$, with $r \ll \min \{ n, m \}$ and $Z$ the space of latent codes.
Identifying this low-dimensional structure is valuable for applications that suffer from the ``curse of dimensionality'', such as uncertainty propagation and Bayesian inference, and is the principle behind dimension reduction techniques such as basis adaptation and active subspaces.

This low-dimensional structure hypothesis is paralleled by the so-called ``manifold hypothesis'' in representation learning, which posits that data in a space of ambient dimension $n$ often lays on a sub-manifold of lower dimension $r \ll n$.
In recent years, there have been significant developments in techniques for unsupervised learning of low-dimensional representations based on the manifold hypothesis~\cite{bengio2014representationlearningreviewnew}.
Of particular note is the family of autoencoder-based techniques~\cite{tschannen2018recent} such as variational autoencoders (VAEs, \cite{kingma2022autoencodingvariationalbayes}) and its variants, which have been successfully used to model complex structures of datasets stemming from various applications.
Recent advances in VAEs aim to improve the ``disentanglement'' properties of VAEs, that is, the capacity of the model to learn representations that maximally separate the various explanatory latent factors (i.e., the components of the latent codes) while discarding as little information as possible.
The $\beta$-VAE formulation~\cite{higgins2017betavae} aims to enhance disentanglement by weighing the Kullback-Leibler (KL) divergence term in the model's variational loss by a factor $\beta$ (which arises naturally for a Gaussian likelihood of the VAE predictions with standard deviation $\sigma^2$, with $\beta \coloneqq \sigma^2$ \cite{pmlr-v139-rybkin21a}).
Other approaches aim to promote disentanglement by encouraging a large mutual information between latent variables and inputs~\cite{kim2018disentangling, chen2018isolating, kumar2018variational} and by promoting independence between latent variable groupings~\cite{lopez2018information, esmaeili2019structured}, among others.

Motivated by these developments in autoencoder-based dimension reduction and disen\-tanglement-promoting techniques, we present a variational encoder-decoder (VED) framework for modeling high-dimensional input-output relations that exhibit the low-dimensional structure hypothesis introduced above.

The proposed framework leverages explicit dimensionality reduction to discover a threshold where latent dimensionality is minimized and information preservation through the encoding/decoding process is maximized.
In this framework, we model the conditional distribution of output data $y \in \mathbb{R}^m$ conditional on input data $x \in \mathbb{R}^n$, $p(y \mid x)$, in terms of parameterized encoder and decoder distribution $p_{\theta}(z \mid x)$ and $q_{\varphi}(y \mid z)$, $z \in Z$, which are formulated using deep learning models. We estimate the VED parameters $\theta, \varphi$ by deriving and minimizing a $\beta$-weighted variational lower bound on the log-conditional distribution of the data, $\log p(y \mid x)$.
Although a similar VED approach was introduced in~\cite{https://doi.org/10.1029/2022MS003130}, the loss function they employ wasn't formally justified.

To promote disentanglement, we equip the proposed variational loss with a term encourages the independence of the latent factors as proposed in~\citet{kumar2018variational}.
This term penalizes the squared entry-wise norm of the difference between the covariance $\operatorname{Cov}_{q_{\varphi}(z)} [z]$ of the so-called ``aggregate encoding distribution'' $q_{\varphi}(z) \coloneqq \mathbb{E}_{p(x)} [ q_{\varphi}(z \mid x) ]$ and the unit covariance, and is weighted by a factor $\lambda$.

To validate the proposed framework, we apply it to model the hydraulic pressure response at observation wells of a two-dimensional groundwater saturated flow model as a function of the model's spatially heterogeneous log-transmissivity field. We pursue non-linear dimensionality reduction techniques as it has been shown that deep-learning feature extraction methods outperform linear methods~\cite{doi:10.1126/science.1127647}. However, we use linear approaches, specifically Canonical Correlation Analysis (CCA, \cite{CanonicalCorrelationandItsRelationshiptoDiscriminantAnalysisandMultipleRegression}) to guide the design of the proposed VED models.
The manuscript is structured as follows:
In Section 2, we describe our methodology for this dimension-reduction task, specifically outlining the problem formulation, loss formulation, model formulation, and computation of informed linear latent representations using CCA. In Section 3, we discuss our results after detailing the dataset preprocessing method, which includes an analysis of our CCA results, a description of the chosen training strategy, an analysis of model performance with respect to latent dimension $r$, and an analysis of our VED's encoding and decoder performance. In Section 4 we discuss our findings in the context of related works, and finally, we present our conclusions in Section 5. 

\section{Variational Encoder-Decoder Formulation}
\label{sec:ved}

\subsection{Variational Loss}
\label{sec:ved:loss}

Let $Z \coloneqq \mathbb{R}^r$ denote the space of latent codes. We propose modeling the data $\{ x \in \mathbb{R}^n, y \in \mathbb{R}^m \}$ employing the parameterized generative model
\begin{equation}
  \label{eq:gen-model}
  p_{\theta}(y, z \mid x) = p_{\theta}(y \mid z) \, p(z \mid x),
\end{equation}
on latent codes $z \in Z$ and output data $y$, conditioned on input data $x$ and parameterized by $\theta$.
Here, $p_{\theta}(y \mid z)$ plays the role of the decoding distribution, randomly mapping latent codes to points in output space.
Marginalizing the generative model over the latent variable leads to the parameterized model $p_{\theta}(y \mid x) = \int p_{\theta}(y \mid z) \, p(z \mid x) \, \mathrm{d} z$, and we aim to identify the parameters $\theta$ for which $p_{\theta}(y \mid x)$ best approximates the true conditional distribution $p(y \mid x)$. We propose employing a variational approach to achieve this goal and formulate a variational lower bound (ELBO) on the log-conditional evidence $\log p_{\theta}(y \mid x)$ by introducing the parameterized variational density $q_{\varphi}(z \mid x)$.

Employing Jensen's inequality leads to the bound
\begin{equation*}
  \begin{split}
    \log p_{\theta}(y \mid x) & \geq \int \log \left [ p_{\theta} (y \mid z) \frac{p(z \mid x)}{q_{\varphi}(z \mid x)} \right ] q_{\varphi}(z \mid x) \, \mathrm{d} z\\
    & = \mathbb{E}_{q_{\varphi}(z \mid x)} \left [ \log p_{\theta}(y \mid z) \right ] - \kldiv{q_{\varphi}(z \mid x)}{p(z \mid x)} \coloneqq \operatorname{ELBO}_{\theta, \varphi}(y \mid x).
  \end{split}
\end{equation*}
Here, we have chosen the variational density $q_{\varphi}(z \mid x)$ to be conditional on $x$ as it approximates the true posterior distribution $p_{\theta}(z \mid x, y)$ of the latent codes given the generative model, and thus we propose using it as the encoding distribution to generate codes from input data. This can be seen by subtracting the ELBO from the log-conditional evidence, which leads to
\begin{equation*}
  \begin{split}
    \log p_{\theta} (y \mid x) - \operatorname{ELBO}_{\theta, \varphi}(y \mid x) &= \int \left \{ \log q_{\varphi}(z \mid x) - \log \left [ \frac{p_{\theta}(y \mid z) p(z \mid x)}{p_{\theta}(y \mid x)} \right ] \right \} q_{\varphi}(z \mid x) \, \mathrm{d} z\\
                                                                                   &= \kldiv{q_{\varphi}(z \mid x)}{p_{\theta}(z \mid x, y)};
  \end{split}
\end{equation*}
from which we can see that the ELBO is equal to the log-conditional evidence when the variational density is equal to the true posterior.
Furthermore, we choose $q_{\varphi}$ to not be conditional on $y$ in order to preserve an encoder-decoder structure in which the distribution of the latent codes is solely conditional on the inputs and the distribution of the outputs is solely conditional on the distribution of the codes.

Based on this formulation, we propose estimating the model parameters $\theta, \varphi$ by maximizing the ELBO averaged over the empirical distribution of the data, $\hat{p}(x, y)$, that is,
\begin{equation}
  \label{eq:ved-loss}
  \max_{\theta, \varphi} \, \mathcal{L}_{\mathrm{VED}}(\theta, \varphi), \quad \mathcal{L}_{\mathrm{VED}} (\theta, \varphi) \coloneqq \mathbb{E}_{\hat{p}(x, y)} \operatorname{ELBO}_{\theta, \varphi}( y \mid x).
\end{equation}
This choice is justified by observing that maximizing the ELBO averaged over the true data-generating distribution $p(x, y)$ is equivalent to minimizing the sum of expected divergences $\mathbb{E}_{p(x, y)} \kldiv{q_{\varphi}(z \mid x)}{p_{\theta}(z \mid x, y)} + \mathbb{E}_{p(x)} \kldiv{p(y \mid x)}{p_{\theta}(y \mid x)}$:
\begin{equation*}
  \begin{split}
    \mathbb{E}_{p(x, y)} \operatorname{ELBO}_{\theta, \varphi}(y \mid x) &= -\mathbb{E}_{p(x, y)} \kldiv{q_{\varphi}(z \mid x)}{p_{\theta}(z \mid x, y)} + \mathbb{E}_{p(x, y)} \left [ \log p_{\theta}(y \mid x) \right ] \\
                                                                         & = -\mathbb{E}_{p(x, y)} \kldiv{q_{\varphi}(z \mid x)}{p_{\theta}(z \mid x, y)}  \\
                                                                         &\quad \quad -\mathbb{E}_{p(x, y)} \left [ \log p(y \mid x) - p_{\theta}(y \mid x) \right ] + \mathbb{E}_{p(x, y)} \left [ \log p(y \mid x ) \right ]\\
                                                                         &= -\mathbb{E}_{p(x, y)} \kldiv{q_{\varphi}(z \mid x)}{p_{\theta}(z \mid x, y)}\\
    &\quad \quad -\mathbb{E}_{p(x)} \kldiv{p(y \mid x)}{p_{\theta}(y \mid x)} + \mathbb{E}_{p(x, y)} \left [ \log p(y \mid x ) \right ],
  \end{split}
\end{equation*}
where we note that the last term on the right-hand side is independent of $\theta, \varphi$.

While the so-called conditional variational autoencoders (CVAEs, \cite{sohn2015learning}) employ a variational loss similar to $\mathcal{L}_{\mathrm{VED}}(\theta, \varphi)$ (see \cite{sohn2015learning}, Eq.~(4)), CVAEs and the proposed VED framework differ in key aspects. Namely, in CVAEs the decoder (referred to as the ``generator network'') is modeled as conditional on both $z$ and $x$, that is, as $p_{\theta}(y \mid z, x)$; furthermore, CVAEs employ parameterized model $q_{\varphi}(z \mid x, y)$ and $p_{\theta}(z \mid x)$ for the variational posterior and the prior, respectively, and separately employ the variational posterior to generate codes at training time and the parameterized prior at testing time.
On the other hand, the VED design aims to achieve explicit dimension reduction, and thus we choose a decoder model conditional only on latent codes and a single encoder model conditional only on the inputs.

The encoding and decoding distributions are parameterized via the multivariate Gaussian models
\begin{align}
  \label{eq:encoder}
  \text{Encoder}: && q_{\varphi}(z \mid x) &\coloneqq \mathcal{N}(z; g_{\varphi}(x), \operatorname{diag} \{ \exp h_{\varphi}(x) \}),\\
  \label{eq:decoder}
  \text{Decoder}: && p_{\theta}(y \mid z) &\coloneqq \mathcal{N}(y; f_{\theta}(z), \sigma^2 I_m),
\end{align}
where $g_{\varphi}(x)$ denotes the encoder mean, $h_{\varphi}(x)$ denotes the encoder diagonal log-variance, $f_{\theta}(z)$ denotes the decoder mean, and $\sigma^2 I_M$ is the decoder covariance.
The encoder model approximates the true posterior of $z$ with a multivariate Gaussian and disregards the posterior correlations.
The decoder model assumes that for a given $z$ the components of the predicted output are normally distributed and mutually independent with equal standard deviation $\sigma$.
Employing these models, the loss in~\eqref{eq:ved-loss} reads
\begin{equation*}
  \mathcal{L}_{\mathrm{VED}}(\theta, \varphi) = \mathbb{E}_{\hat{p}(x, y)} \left \{ \frac{1}{2 \sigma^2} \mathbb{E}_{z \sim q_{\varphi}(z \mid x)} \left \| y - f_{\theta} (z) \right \|^2_2 + \kldiv{q_{\varphi}(z|x)}{\mathcal{N}(0, I_r)} \right \},
\end{equation*}
where we have chosen for simplicity the prior $p(z \mid x) \sim \mathcal{N}(0, I_r)$.
It can be seen that the loss consists of the sum of a mean (over the data and the stochastic output of the decoder) squared error loss and a mean (over the data) KL divergence loss.
This KL divergence is between two multivariate Gaussian distributions, and so it can be computed in closed form in terms of $g_{\theta}$ and $h_{\theta}$~\cite{doersch2021tutorialvariationalautoencoders}.
The mean squared error loss is computed via the so-called ``reparameterization trick'' as
\begin{equation}
  \label{eq:mse}
  \operatorname{MSE}(\theta, \varphi) \coloneqq \mathbb{E}_{\hat{p}(x, y)} \left \{ \mathbb{E}_{\epsilon \sim \mathcal{N}(0, I_m)} \left \| y - f_{\theta} \left ( g_{\varphi}(x) + \epsilon \odot \exp \tfrac{1}{2} h_{\varphi}(x) \right ) \right \|^2_2 \right \},
\end{equation}
where $\odot$ denotes element-wise (Hadamard) product and $\epsilon \sim \mathcal{N}(0, I_m)$ is a vector of standard normal fluctuations. In practice, the expectation over $\epsilon$ is computed using a single draw of $\epsilon$ per training data sample.
Furthermore, we define the KL divergence loss as
\begin{equation}
  \label{eq:kld}
  \operatorname{KLD}(\varphi) \coloneqq \mathbb{E}_{\hat{p}(x)} \left [ \kldiv{q_{\varphi}(z|x)}{p(z \mid x)} \right ].
\end{equation}
Substituting these definitions above we obtain the $\beta$-weighted loss
\begin{equation}
  \label{eq:loss}
  \mathcal{L}^{\beta}_{\mathrm{VED}}(\theta, \varphi) \coloneqq \tfrac{1}{2} \operatorname{MSE}(\theta, \varphi) + \beta \operatorname{KLD}(\varphi),
\end{equation}
where the parameter $\beta \coloneqq \sigma^2$ controls the weight of the KLD loss relative to the MSE loss.

\subsection{Disentanglement-Promoting Regularization}
\label{sec:ved:disentanglement}

\citet{pmlr-v139-rybkin21a} found that the $\beta$ parameter can be tuned to improve model performance by balancing reconstruction and generating capacity. Preliminary testing for a range of $\beta$ values ($0$, $0.001$, $0.01$, $0.1$, $0.5$, and $1.0$) indicated that solely tuning $\beta$ led to too strong a compromise between reconstruction and generating capacity, with high reconstruction capacity and low generating capacity for low $\beta$, and the opposite for high $\beta$.
We hypothesize that we can improve this balance by enhancing the disentanglement of latent features, and following \citet{mathieu2019disentanglingdisentanglementvariationalautoencoders} we propose equipping the $\beta$-VED loss \eqref{eq:loss} with an additional disentanglement-promoting regularization term.
Let $q_{\varphi}(z) \coloneqq \int q_{\varphi}(z \mid x) \, \hat{p}(x) \, \mathrm{d} x$ denote the so-called ``aggregate encoding distribution'', defined as the distribution of latent codes given by the encoder marginalized over the empirical distribution of the input data $\hat{p}(x)$.
In this work we employ the DIP-VAE-II regularizing term $\operatorname{COV}(\varphi)$ proposed in \citet{kumar2018variational}, which encourages disentanglement by penalizing the squared entry-wise norm of the difference between $\operatorname{Cov}_{q_{\varphi}(z)} [z]$ and the unit covariance, that is,
\begin{equation}
  \label{eq:cov_loss}
  \operatorname{COV}(\varphi) := \sum\nolimits_{i \neq j} \left \{ \text{Cov}_{q_\varphi(z)}[z] \right \}_{ij}^2 + \sum\nolimits_i \left ( \left \{ \operatorname{Cov}_{q_\varphi(z)}[z] \right \}_{ii} - 1 \right )^2.
\end{equation}
$\operatorname{Cov}_{q_{\varphi}(z)} [z]$ for the decoder \eqref{eq:encoder} can be computed employing the law of total variance as
\begin{equation*}
  \begin{split}
    \operatorname{Cov}_{q_{\varphi}(z)} [z] &= \mathbb{E}_{\hat{p}(x)} \operatorname{Cov}_{q_{\varphi}(z \mid x)} [z] + \operatorname{Cov}_{\hat{p}(x)} \left \{ \mathbb{E}_{q_{\varphi}(z \mid x)} [z] \right \}\\
    &= \operatorname{diag} \left \{ \mathbb{E}_{\hat{p}(x)} \left [ \exp h_{\varphi}(x) \right ] \right \} + \operatorname{Cov}_{\hat{p}(x)} g_{\varphi}(x).
  \end{split}
\end{equation*}
Another interpretation of the $\operatorname{COV}(\varphi)$ is that, by encouraging the aggregate encoding distribution to be close to the prior $p(z \mid x) = \mathcal{N}(0, I_r)$, it leads to the distribution of VED-generated outputs,
\begin{equation*}
  p_{\theta, \varphi}(y) = \int p_{\theta}(y \mid z) \left ( \int q_{\varphi}(z \mid x) \, p(x) \, \mathrm{d} x \right) \mathrm{d} z = \int p_{\theta} (y \mid z) \, q_{\varphi}(z) \, \mathrm{d} z,
\end{equation*}
more closely resembling the distribution of outputs generated employing the generative model \eqref{eq:gen-model}, $p_{\theta}(y) = \int p_{\theta}(y \mid z)$, that is, it enhances the generative capacity of the VED model.

Although each of the two terms in \eqref{eq:cov_loss} can be weighted separately, in this work we weight them equally.
The final loss function takes the form
\begin{equation}
  \label{eq:loss_final}
  \mathcal{L}^{\beta, \lambda}(\theta, \varphi) \coloneqq \tfrac{1}{2} \operatorname{MSE}(\theta, \varphi) + \beta \operatorname{KLD}(\theta) + \lambda \operatorname{COV}(\varphi)
\end{equation}
where we have introduced the penalty weight $\lambda > 0$.
While the weight $\lambda$ is not formally motivated as is $\beta$, and the introduction of the $\operatorname{COV}(\varphi)$ penalty precludes the interpretation of the loss as a (negative) averaged ELBO, we find in practice that the introduction of this penalty improves both the reconstruction and generative capacities of the VED model.

\section{Low-Dimensional Surrogate Modeling Application}

To validate the proposed VED framework, we apply it to modeling the observable hydraulic pressure response of a groundwater flow model of the Hanford Site as a function of the model's log-hydraulic transmissivity field.
The numerical flow model employs a cell-centered finite volume (FV) discretization and assumes two-dimensional, saturated flow, as described in \cite{yeung2024gaussian}.
We note that this model is a simplification of the true physics at the Hanford Site, which is characterized by transient unsaturated flow. 
The observable response corresponds to the stationary pressure value at $323$ observation wells, and the input corresponds to the $1475$ discrete log-transmissivity values corresponding to the cells of the FV model.
The dataset consists of $20,000$ draws of the log-transmissivity and the corresponding pressure response.
The log-transmissivity fields were drawn by sampling the $1,000$-dimensional Kosambi-Karhunen-Loève expansion (KKLE) of a Gaussian process model trained using data from \citet{cole-2001-transient}'s 2001 Hanford Site calibration study.

Notably, this data represents a relationship between high-dimensional inputs ($1475$-dimensional, with an intrinsic dimension of $1,000$ stemming from the KKLE) and high-dimensional outputs ($323$ output features). Thus, we attempt to discover the low-dimensional structure of this high-dimensional input-output relationship using the proposed VED framework.

\subsection{Preliminary Data Analysis and Pre-Processing}

We apply canonical correlation analysis (CCA) to this dataset to analyze its low-dimensional structure. Specifically, we employ the latent dimension of the CCA-based linear encoding of the input-output relationship to guide our choice of $r$ for the VED experiments. The CCA encoding is constructed by solving the generalized eigenvalue problem \cite{CanonicalCorrelationandItsRelationshiptoDiscriminantAnalysisandMultipleRegression}
\begin{equation*}
  S_{XY} S_{YY}^{-1} S_{YX} a = s^2 (S_{XX} + \varepsilon I_n),
\end{equation*}
where $S_{XX}$ and $S_{YY}$ are the input and output sample covariances, respectively, and $S_{XY}$, $S_{YX} = S_{XY}^{\top}$ are the input-output and output-input sample cross-covariances, respectively.

\begin{figure}[!htb]
    \centering
    \includegraphics[width=3in]{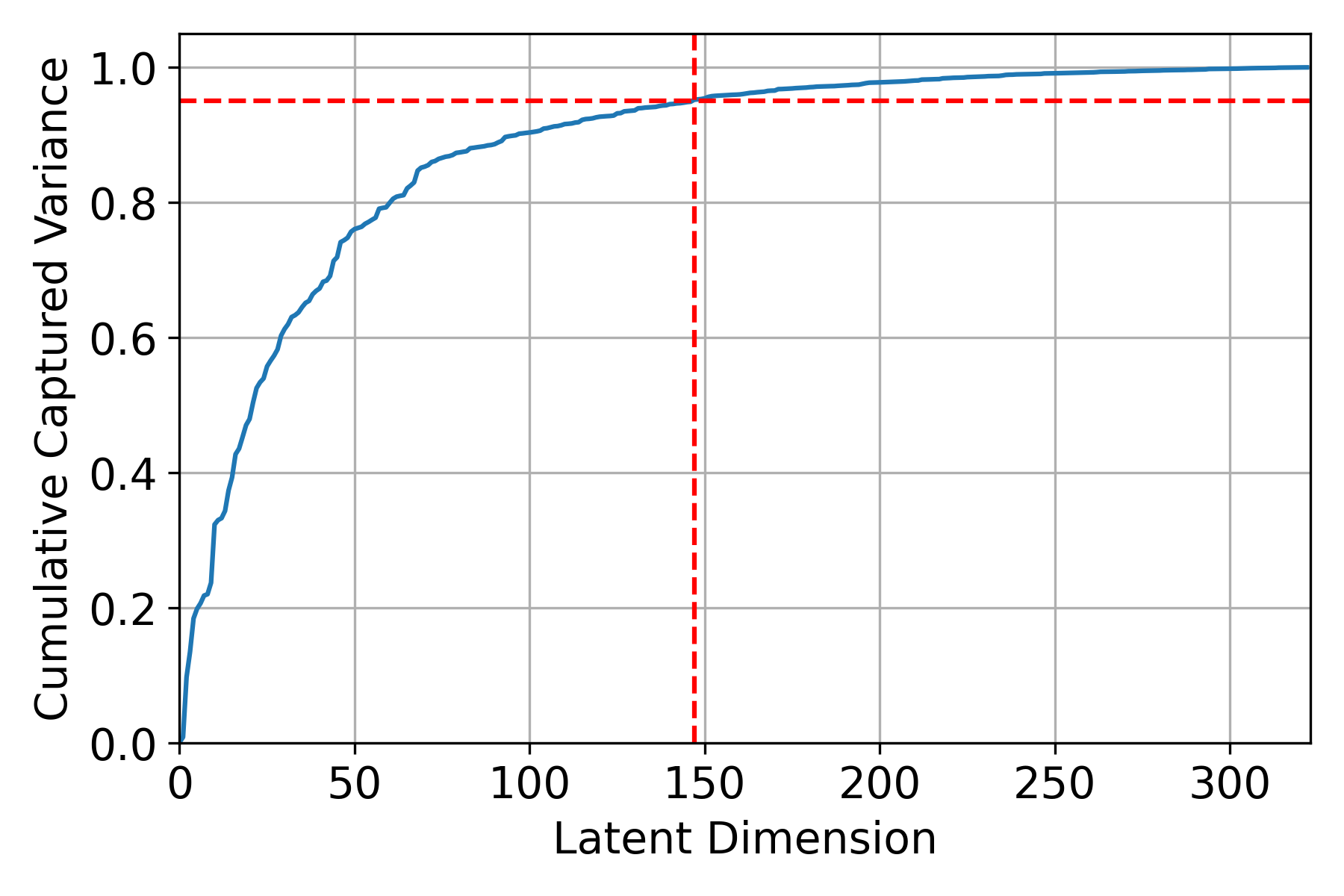}
    \caption{CCA truncated explained output variance as a function of the latent dimension. The red dashed line shows where we capture $95\%$ of the CCA explained variance ($r = 147$).}
    \label{fig:cca_data}
\end{figure}

The solution of the previous generalized eigenvalue problem results in $k \coloneqq \min \{n, m \}$ pairs of eigenvalues and (column) eigenvectors $\{ s_i, a_i \}$, which we organize in descending order with respect to $s^2$ so that $s^2_1 \geq s^2_2 \geq \cdots \geq s^2_k$. Furthermore, we organize the eigenvectors into the matrix $A = [a_1, \dots, a_k]$, and define $C \coloneqq S_{YX} A$.
The output variance explained by the linear encoding consisting of the first $i$ eigenpairs is given by
\begin{equation*}
  \text{CEV}_i = \operatorname{trace} \, ( C_{,1:i} C^{\top}_{,1:i} ),
\end{equation*}
where $C_{,1:i}$ denotes the fist $i$ columns of $C$. Figure~\ref{fig:cca_data} shows $\text{CEV}_i$ normalized by the total explained variance $\text{CEV}_k$. It can be seen that $147$ latent features are sufficient to capture $95\%$ of the total explained variance (which is less than the total output variance; in fact, the CCA-based linear encoding captures at most $\sim 70\%$ of the total variance, which indicates that linear encoding is sub-optimal for this dataset).
Informed by the CCA results, we expect the non-linear VED dimensionality reduction to perform better than this linear estimate, as non-linear deep learning-based encoding techniques often outperform linear methods \cite{doi:10.1126/science.1127647}. Thus, we employ the choices of latent dimension $r = 50$, $100$, $150$, and $200$ to investigate whether the VED model can discover an encoding for $r < 147$ without significant loss of information.

The Hanford site dataset's log-transmissivity input consists of $1475$ corresponding to the FV model's irregular quadrilateral grid 

\begin{figure}[!htb]
    \centering
    \includegraphics[width=3.5in]{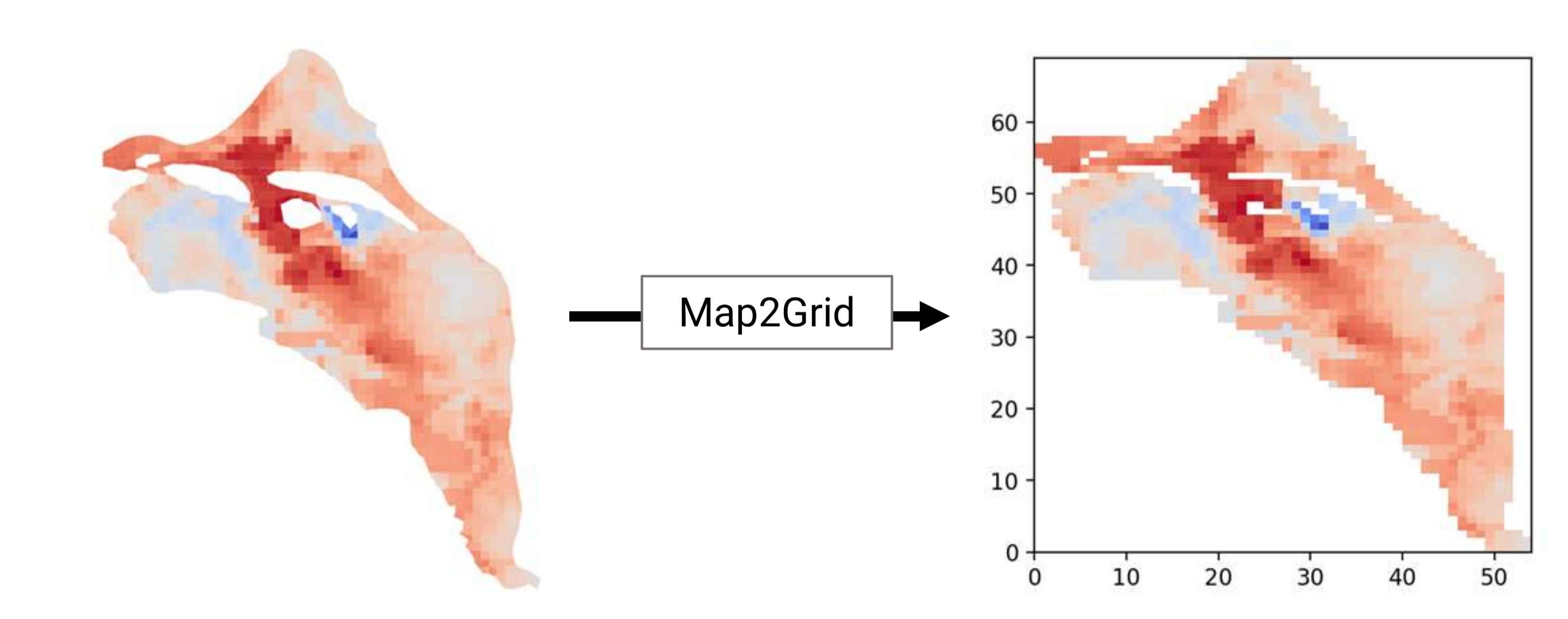}
    \caption{\texttt{Map2Grid} data mapping from an irregular grid to a Cartesian grid.}
    \label{fig:true_data}
\end{figure}

The input data, which corresponds to cell-centered log-transmissivity values over the FV model's irregular quadrilateral grid (Figure~\ref{fig:true_data}), is not amenable to modeling via traditional convolutional models as such models require consistent and uniform receptive fields for feature extraction.
Thus, we developed a \texttt{Map2Grid} function that maps each $1475$-dimensional input to a $69 \times 54$ Cartesian grid, which allows for this preprocessed data to be used as input for convolutional models. This transformation is shown in Figure~\ref{fig:true_data}.
We account for empty cells by calculating the loss over each batch using a pre-computed mask, allowing for the empty cells to not interfere with model reconstructions.
An alternative approach is to define convolutions over the irregular grid using graph convolutional layers~\cite{kipf2017gcnn}; however, we choose to employ the \texttt{Map2Grid}-based procedure described above as it is more computationally straightforward. The use of graph convolutions for VEDs will be explored in future work.

\subsection{Model Formulation}

We initially formulate a set of baseline VED models with convolutional layers in the encoder, finding that this structure outperforms fully connected encoder-decoder setups. This aligns with the general consensus on CNNs \cite{726791} which, due to the added feature extraction and hierarchical feature learning from stacked convolutions \cite{zeiler2013visualizingunderstandingconvolutionalnetworks, Masci_feature_extraction, Alexnet}, provide increased learning capabilities, specifically for high-dimensional data. Taking inspiration from benchmark pure CNN setups \cite{Alexnet, 726791,zeiler2013visualizingunderstandingconvolutionalnetworks}, we find that approaches using aggressive striding and large feature expansion with decreased image size over multiple convolutions work best for balancing reconstruction accuracy and KLD regularization of our loss. We find an optimal setup by applying striding five times throughout the encoder, which decreases the image size from our (69, 54) input to a (2, 1) size after applying convolutions, and we additionally increase the number of features as $1 \rightarrow 16 \rightarrow 32 \rightarrow 64 \rightarrow 128 \rightarrow 256$ across our convolutions. 

\begin{figure}[!htb]
    \begin{subfigure}[b]{0.49\textwidth}  
        \centering
        \includegraphics[width=\textwidth]{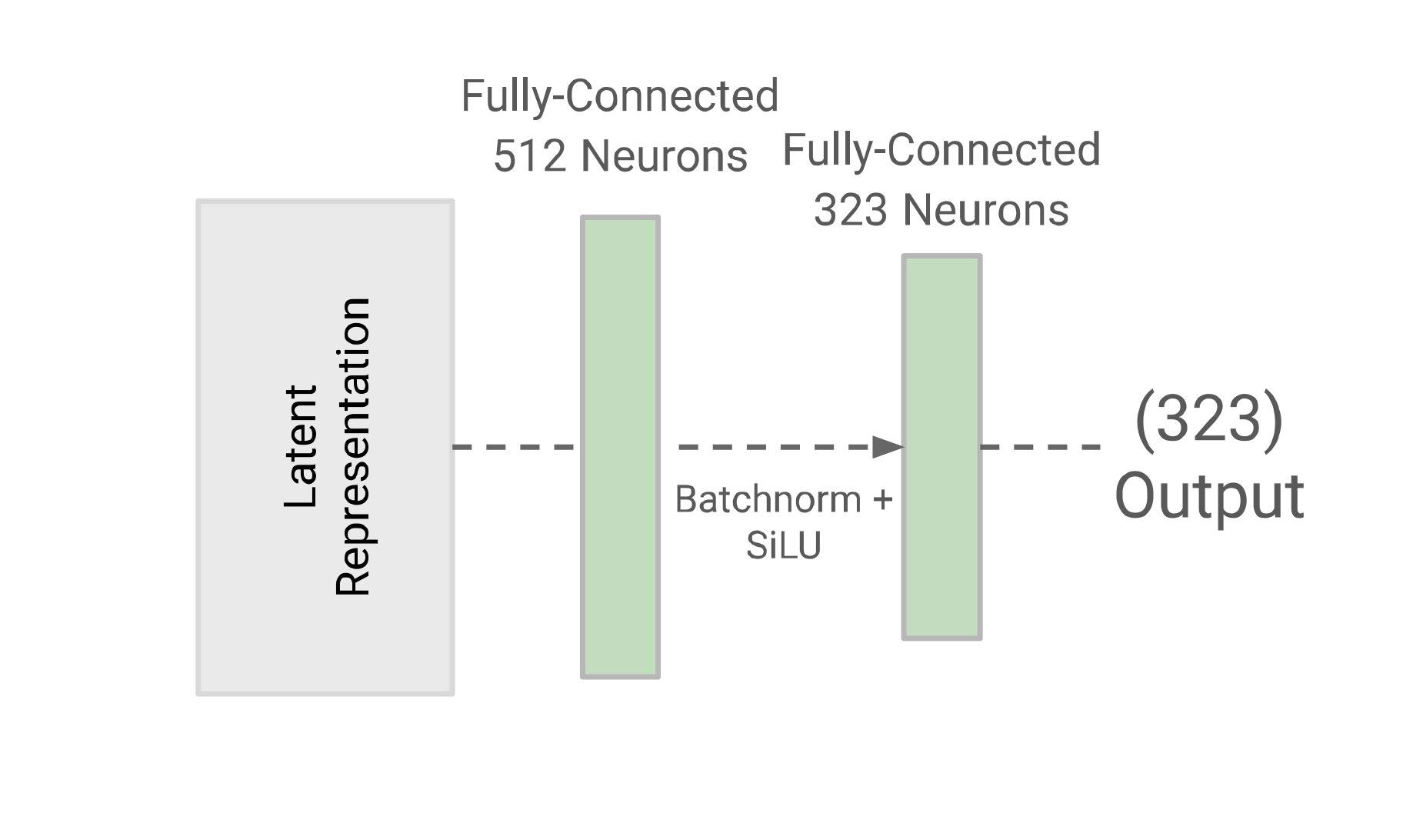}
        \caption{}
        \label{fig:decoder_architecture}
    \end{subfigure}
    \hfill
    \begin{subfigure}[b]{0.49\textwidth}
        \centering
        \includegraphics[width=\textwidth]{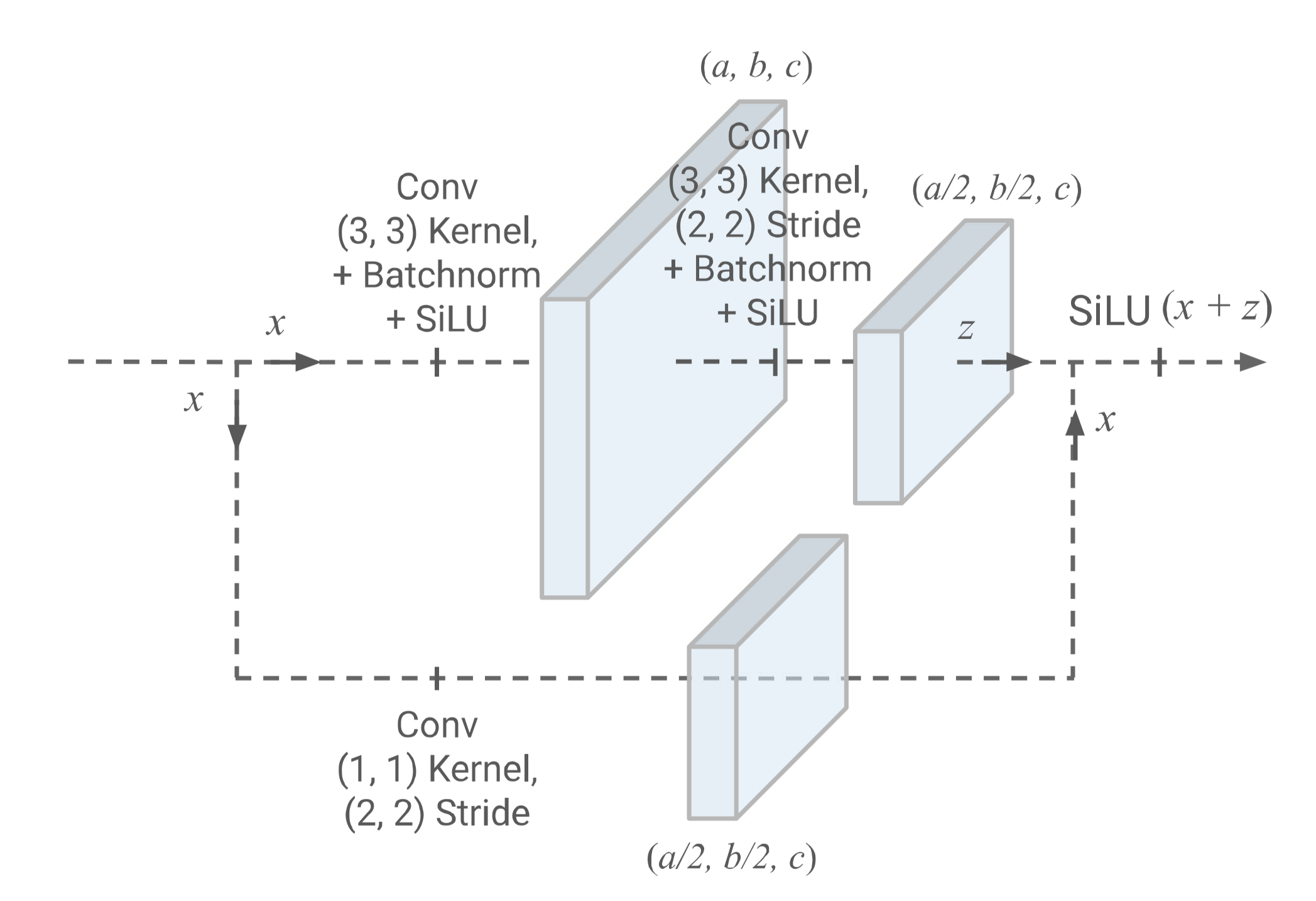}
        \caption{}
        \label{fig:res_block}
    \end{subfigure}
    \caption{(a) Decoder architecture. (b) Residual block architecture.}
    \label{fig:additional_architecture}
\end{figure}

Notably, we also find through experimenting with different encoder and decoder depths that combining deep encoders with shallow decoders leads to good performance. Across all of our tests, we find that the best-performing decoder structure simply employs two fully connected layers, with one layer increasing the dimensionality of our latent representation, from $r$ to size $512$, followed by batch normalization, followed by an output layer that transforms our data to the desired $323$ output features (Figure~\ref{fig:decoder_architecture}). This deeper-encoder shallow-decoder structure is supported by previous works (e.g., \cite{kasai2021deepencodershallowdecoder}) which find that sufficiently deep encoders can utilize simple decoders to achieve good performance.

\begin{figure}[!htb]
    \centering
    \includegraphics[width=1\textwidth]{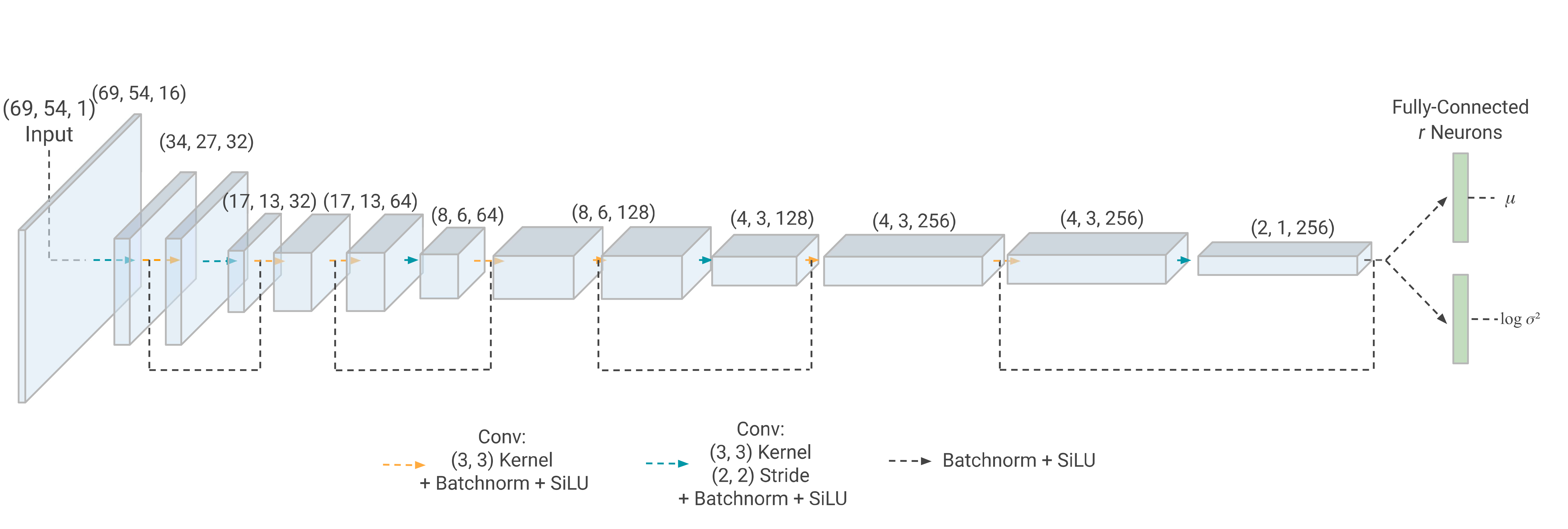}
    \caption{Encoder architecture. Blocks represent data shape and arrows represent transformations between data blocks.}
    \label{fig:architecture}
\end{figure}

We find the largest improvement in performance by switching from simple convolutional layers to residual blocks \cite{he2015deepresiduallearningimage}. We use a residual block structure with two convolutional blocks each followed by one batch normalization layer \cite{batchnorm}, with striding of 2 being applied at the first convolution. Thus, the residual block halves our image size but maintains the number of features, and we increase the number of features in the processed encoding with a convolutional layer before each residual block (Figure~\ref{fig:res_block}). The best-performing models employing this structure include four residual blocks, and 13 total convolutional layers in our encoder, where the final (2, 1, 256) dimensional output is channeled separately into two fully-connected layers with $r$ neurons that output the encoder mean $g_{\varphi}$ and log-variance $h_{\varphi}$ (Figure~\ref{fig:architecture}).


\section{Numerical Experiments}

\subsection{Training Strategy}

We employ the Adam~\cite{kingma2017adammethodstochasticoptimization} optimizer to minimize the loss~\eqref{eq:loss_final} and find through experimentation that a cosine-decay learning rate scheduler with a batch size of $100$ samples results in the best performance. Additionally, we use gradient clipping \cite{pascanu2013difficulty} to mitigate the problem of exploding gradients by limiting the magnitude of the gradient updates during backpropagation.
We considered step-wise, linear, and cyclic annealing schedules for the $\beta$ and $\lambda$ parameters for which the values of the parameters change in step-wise, linear, and cyclic schedules according to training epochs.
Inspired by other work in VAEs that have found success in using annealing schedules for the $\beta$ parameter during training \cite{https://doi.org/10.1029/2022MS003130, jacobsen2022disentangling}, we performed preliminary experiments with step-wise, linear and cyclic schedules for both the $\beta$ and $\lambda$ parameters. Nevertheless, we find that these schedules do not perform as well as constant-parameter schedules for our problem.

Therefore, we perform experiments for the parameter values of $r \in \{50, 100, 150, 200\}$,  $\beta \in \{0, 0.01, 0.1 \}$, and $\lambda \in \{0, 0.01, 0.1 \} $.
We train the VED model for these parameter values and using training subsets of $5,000$, $10,000$ and $15,000$ samples over $100$ epochs, and for all experiments we report test metrics computed using a test subset of $5,000$ samples.
Results for the experiments with $5,000$ training samples are reported in this section, while results for the $10,000$ and $15,000$ training dataset sizes are presented in Appendix~\ref{sec:10k-15k}.

\subsection{Model Performance w.r.t. Latent Dimension}

\begin{figure}[!htb]
    \centering
    \begin{subfigure}[b]{0.9\textwidth}
        \centering
        \includegraphics[width=\textwidth]{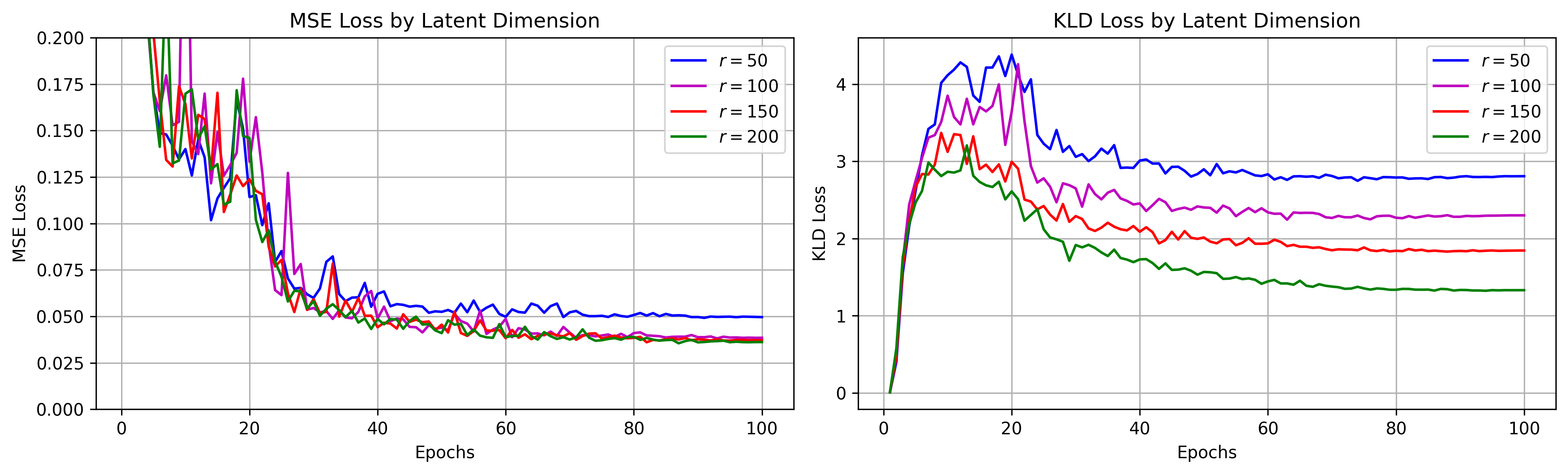}
        \caption{$\beta = 0.01, \lambda = 0.01$, varying $r$.}
        \label{fig:loss-vary-r}
    \end{subfigure}\\
    \begin{subfigure}[b]{0.9\textwidth}
        \centering
        \includegraphics[width=\textwidth]{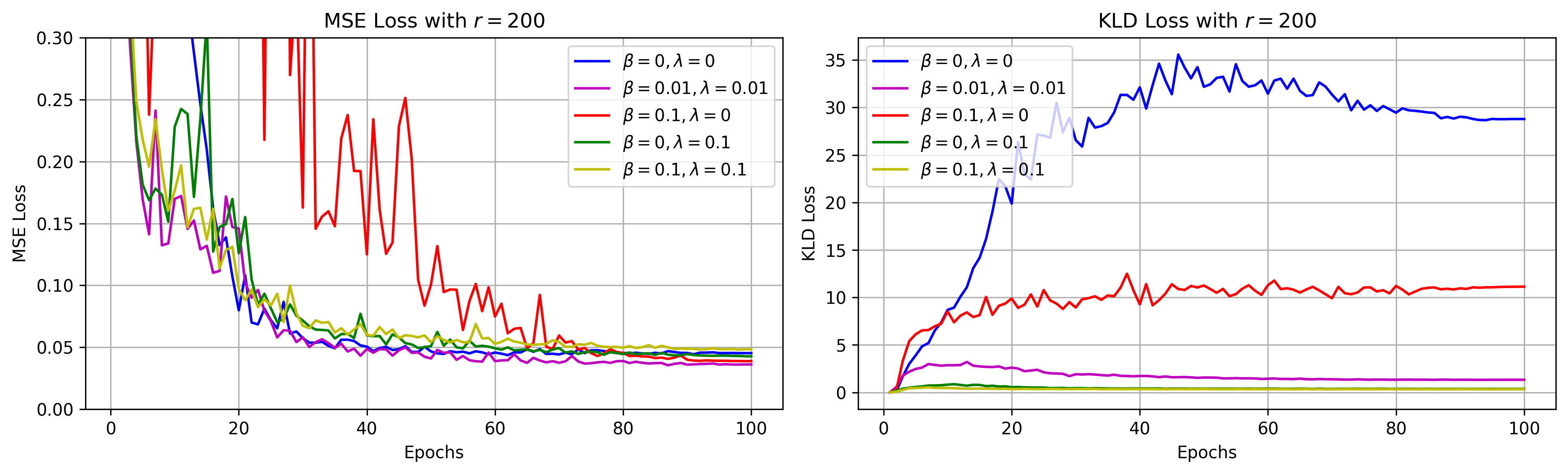}
        \caption{$r = 200$, varying $\beta$ and $\lambda$.}
        \label{fig:loss-fixed-r}
    \end{subfigure}
    \caption{Loss histories over 100 training epochs.}
    \label{fig:loss}
\end{figure}

For each experiment, we compute the MSE and KLD metrics for the test data after each epoch and store the set of parameters $\theta, \varphi$ that result in the best test MSE. Table~\ref{tab:my_label} presents the best MSE and the corresponding KLD, and we highlight in bold the best MSE and corresponding KLD for each choice of $r$.
The reported MSE is calculated for normalized test data (normalized with respect to the mean and standard deviation of the training data) and averaged over the number of output features, $m = 323$. Similarly, we report the KLD averaged over the number of latent features $r$. We also 
present in Figure~\ref{fig:loss} the training histories of the MSE and KLD tests.

\begin{figure}[!htb]
    \centering
    \begin{subfigure}[b]{\textwidth}
        \centering
        \includegraphics[width=0.9\textwidth]{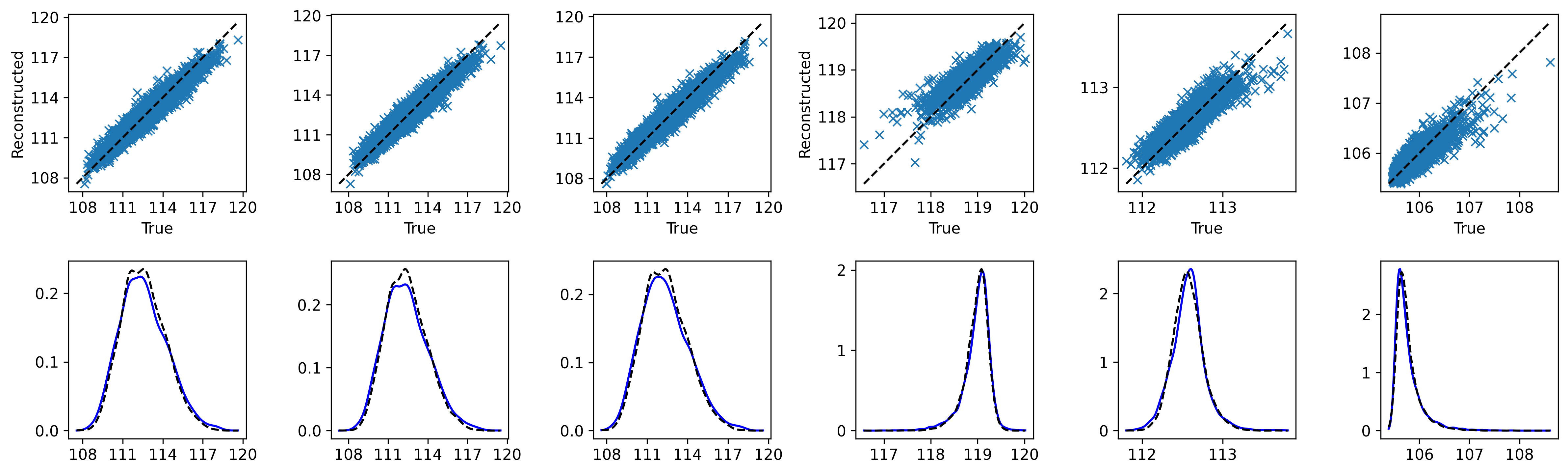}
        \caption{$r$ = 50}
    \end{subfigure}
    \begin{subfigure}[b]{\textwidth}
        \centering
        \includegraphics[width=0.9\textwidth]{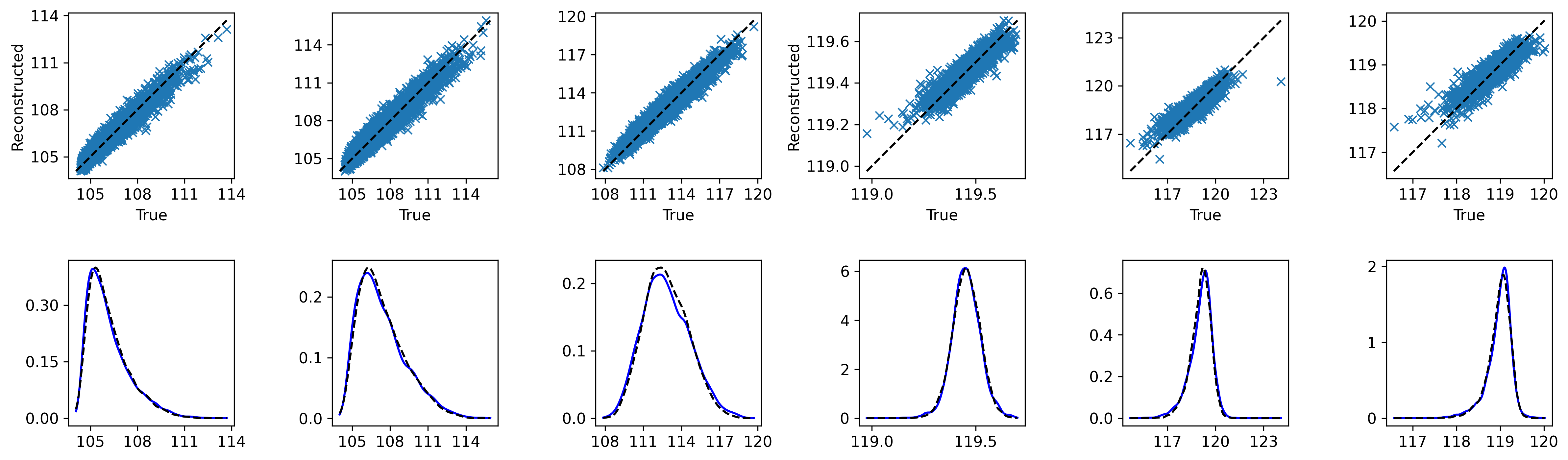}
        \caption{$r$ = 100}
    \end{subfigure}
    \begin{subfigure}[b]{\textwidth}
        \centering
        \includegraphics[width=0.9\textwidth]{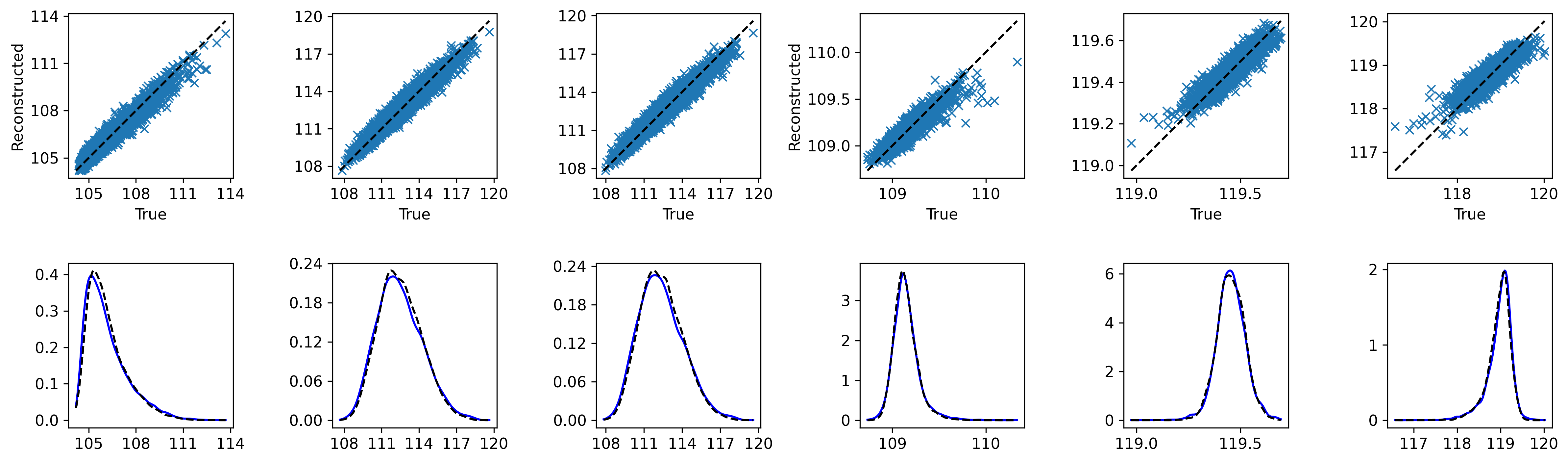}
        \caption{$r$ = 200}
    \end{subfigure}
    \caption{Reconstruction accuracy of best 3 indices (left) and worst 3 indices (right) out of reconstructed 323 output features, for $\beta = 0.01, \lambda = 0.01$, sorted by RMSE. Densities computed using Gaussian kernel density estimation. Blue densities indicate test data, and dashed black densities indicate reconstructions.}
    \label{fig:ldims}
\end{figure}

Figure~\ref{fig:loss-vary-r} shows the training histories for all values of $r$ considered and fixed $\beta = 0.01, \lambda = 0.01$, and Figure~\ref{fig:loss-fixed-r} shows the histories for fixed $r = 200$ and various choices of $\beta, \lambda$. Furthermore, we present in Figure~\ref{fig:ldims} a comparison between true and reconstructed test data for the 3 output features with the lowest MSE and the 3 features with the largest MSE calculated for fixed $\beta = 0.01$, $\lambda = 0.01$ and various values of $r$ ($50$, $100$, and $200$).

\begin{table}[!htb]
    \centering
    \begin{tblr}{
        colspec = {l l | l  l | l  l  | l l | l l |},
  cell{1}{1} = {c = 2,r = 2}{c,m},
  cell{1}{3} = {c = 2}{halign = c},
  cell{1}{5} = {c = 2}{halign = c},
  cell{1}{7} = {c = 2}{halign = c},
  cell{3}{1} = {r = 3}{valign = m},
  cell{6}{1} = {r = 3}{valign = m},
  cell{9}{1} = {r = 3}{valign = m},
  cell{12}{1} = {r = 3}{valign = m},
  hline{3}   = {black},
  hline{4}   = {black, dotted},
  hline{5}   = {black, dotted},
  vline{4}   = {black, dotted},
  hline{6}   = {black},
  hline{7}   = {black, dotted},
  hline{8}   = {black, dotted},
  vline{6}   = {black, dotted},
  hline{9}   = {black},
  hline{10}   = {black, dotted},
  hline{11}   = {black, dotted},
  vline{8}   = {black, dotted},
  hline{12}   = {black},
  hline{13}   = {black, dotted},
  hline{14}   = {black, dotted},
  hline{15}   = {black}
  }
  &  &  $\lambda = 0$ & & $\lambda = 0.01$ & & $\lambda = 0.1$ & \\
  &  &  MSE & KLD & MSE & KLD & MSE & KLD\\
  $r = 50$ &  $\beta = 0$ &  0.0501 & 57.044 & 0.0505 & 7.6910 & \textbf{0.0484} & \textbf{7.4875}\\
  &  $\beta = 0.01$ &  0.0531 & 2.7745 & 0.0496 & 2.7896 & 0.0485 & 2.8197\\
  &  $\beta = 0.1$ &  0.0496 & 1.4661 & 0.0501 & 1.4612 & 0.0504 & 1.4207\\
  $r = 100$ &  $\beta = 0$ &  0.0438 & 39.380 & 0.0396 & 8.0034 & \textbf{0.0383} & \textbf{8.5265}\\
  &  $\beta = 0.01$ &  0.0416 & 2.2356 & 0.0384 & 2.2990 & 0.0388 & 2.2413\\
  &  $\beta = 0.1$ &  0.0442 & 0.7756 & 0.0449 & 0.7640 & 0.0474 & 0.7141\\
  $r = 150$ &  $\beta = 0$ &  0.0430 & 36.246 & 0.0386 & 8.1343 & 0.0384 & 10.463\\
  &  $\beta = 0.01$ &  0.0385 & 1.7932 & 0.0368 & 1.8289 & \textbf{0.0367} & \textbf{1.4570}\\
  &  $\beta = 0.1$ &  0.0434 & 0.5111 & 0.0442 & 0.5036 & 0.0473 & 0.4652\\
  $r = 200$ &  $\beta = 0$ & 0.0422 & 33.072 & 0.0386 & 7.9668 & 0.0385 & 11.266\\
  &  $\beta = 0.01$ & 0.0433 & 1.4166 & 0.0359 & 1.3225 & \textbf{0.0350} & \textbf{1.0909}\\
  &  $\beta = 0.1$ & 0.0427 & 0.3755 & 0.0441 & 0.3690 & 0.0486 & 0.3342\\
    \end{tblr}
    \caption{Reconstruction and KLD test loss values for $\beta \in \{0, 0.01, 0.1\}, \lambda \in \{0, 0.01, 0.1\}, r \in \{50, 100, 150, 200\}$ on training dataset with 5,000 samples.}
    \label{tab:my_label}
\end{table}

It can be seen from Table~\ref{tab:my_label} and Figure~\ref{fig:loss} that while increasing the latent dimension from $r = 50$ to $100$ decreases MSE by $\sim 22\%$, further increasing $r$ from $100$ to $200$ decreases MSE more modestly, about $\sim 7\%$. This indicates that $r$ values of $\sim 100$ are sufficient to represent the input-output relationship for this dataset.
This can be seen clearly in Figure~\ref{fig:loss-vary-r}, which shows a small reduction in reconstruction accuracy when going from $r = 200$ to $100$ and a modest reduction when going from $r = 100$ to $r = 50$.
Additionally, Figure~\ref{fig:ldims} shows that the VED model is capable of modeling the input-output relationship and the distribution of the output data even at $r = 50$.
While there is an improvement in reconstruction accuracy for higher latent dimensions, the reconstruction capacity of the model with $r = 50$ is still very good, with a test MSE (averaged across output features) of around $0.05$.

We now proceed to analyze the effect of the KL and disentanglement-promoting regularization terms on reconstruction performance.
Setting $\beta = 0$ yields the best reconstruction performance for small $r$. However, as we increase $r$ we find the reconstruction performance to occur for non-zero $\beta$, which indicates that KL regularization can improve reconstruction performance when paired with disentanglement-promoting regularization.
Notably, this does not occur for larger training datasets (Tables~\ref{tab:10k-table} and~\ref{tab:15k-table}), for which we find that $\beta = 0$ and non-zero $\lambda$ leads to the best performance across all values of $r$ studied, indicating that, for larger datasets, disentanglement-promoting regularization is more important than KL regularization for accuracy.
At $\beta = 0.1$, we start to see losses in reconstruction performance that become more pronounced as $\beta$ increases further. 

With respect to $\lambda$, we find that the smallest MSE values for each choice of $r$ occur with $\lambda$, indicating that the disentanglement-promoting penalty actually has a favorable effect on reconstruction performance.
However, in the case of large $\beta$, e.g. $\beta = 0.1$, increasing $\lambda$ actually decreases model accuracy. Ultimately, we find that for the constrained choices $\lambda \in \{0, 0.01, 0.1\}$, disentanglement-promoting regularization term does not significantly decrease reconstruction performance as is often the case for other regularization approaches yielding lower reconstruction quality in the context of VAEs \cite{tschannen2018recent}, and in fact can contribute to improving accuracy in many cases.

Finally, it can be seen from Table~\ref{tab:my_label} that small, non-zero values for $\beta$ and $\lambda$ lead to the best balance between the MSE and KLD metrics for all values of $r$ considered.

Tables~\ref{tab:10k-table} and~\ref{tab:15k-table} show that although the lowest values of test MSE for the experiments with larger training datasets are found for $\beta = 0$ and non-zero $\lambda$ across all values of $r$ tested, a small value of $\beta$ results in significant reductions of the KLD metric without significantly degrading reconstruction performance.
Together with the results presented in Section~\ref{sec:res:decoding}, this indicates that the KL and disentanglement-promoting regularizations both contribute to balancing the reconstruction and generative capacities of the VED model.

\subsection{Encoding and Decoding Performance}
We also investigate how training with $\beta$ and $\lambda$ parameters (in constrained ranges as to not affect model reconstruction performance) affected the model's generative capabilities, which we tested by isolating the encoder and the decoder. 

\subsubsection{Decoded Gaussian Noise}
\label{sec:res:decoding}

\begin{figure}[!htb]
    \centering
    \begin{subfigure}[b]{\textwidth}
        \centering
        \includegraphics[width=0.8\textwidth]{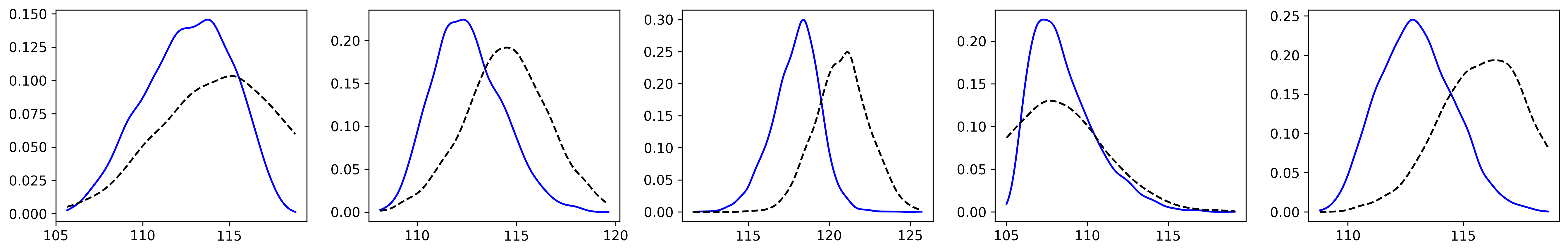}
        \caption{$\beta = 0, \lambda = 0.1$}
    \end{subfigure}
    \begin{subfigure}[b]{\textwidth}
        \centering
        \includegraphics[width=0.8\textwidth]{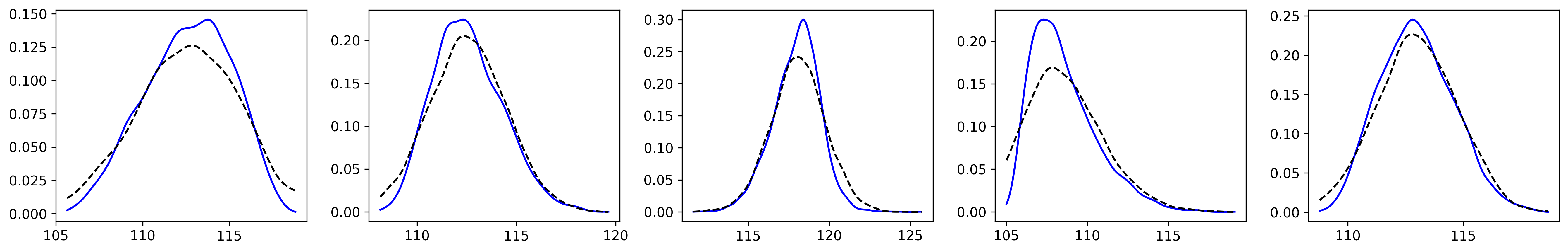}
        \caption{$\beta = 0.01, \lambda = 0.1$}
    \end{subfigure}
    \begin{subfigure}[b]{\textwidth}
        \centering
        \includegraphics[width=0.8\textwidth]{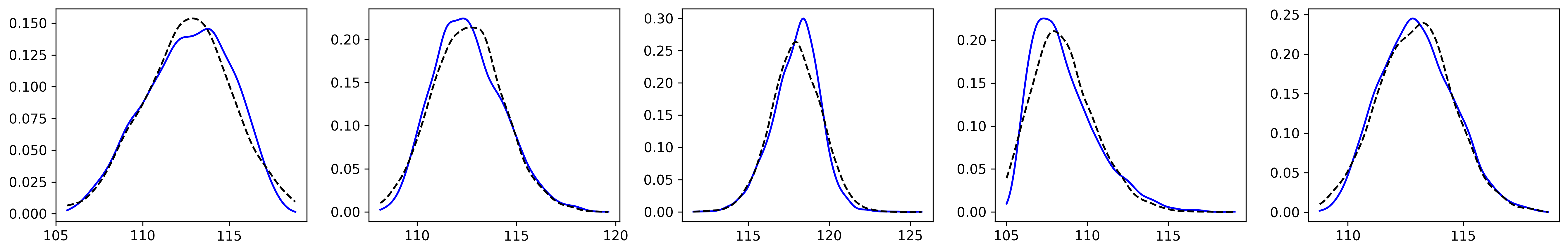}
        \caption{$\beta = 0.1, \lambda = 0.1$}
    \end{subfigure}
    \begin{subfigure}[b]{\textwidth}
        \centering
        \includegraphics[width=0.8\textwidth]{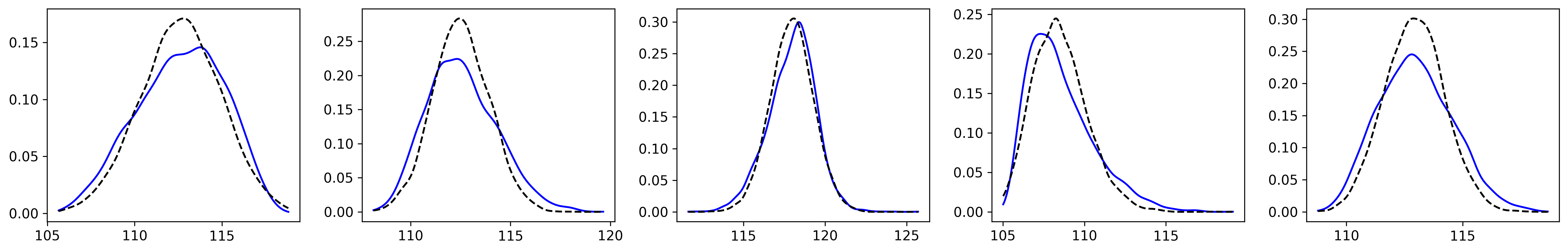}
        \caption{$\beta = 0.1, \lambda = 0.01$}
    \end{subfigure}
    \begin{subfigure}[b]{\textwidth}
        \centering
        \includegraphics[width=0.8\textwidth]{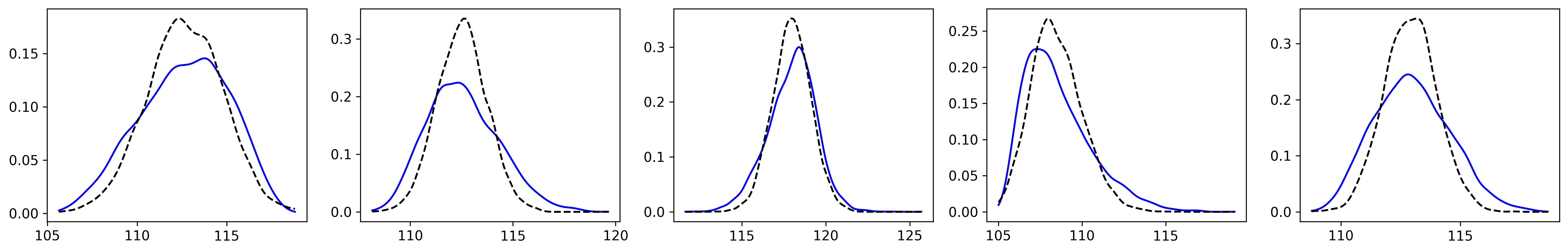}
        \caption{$\beta = 0.
        1, \lambda = 0$}
    \end{subfigure}
    \caption{Synthetic data generated by decoding Gaussian noise with $r = 200$. Densities computed using Gaussian kernel density estimation. Blue densities indicate test data, and dashed black densities indicate reconstructions.}
    \label{fig:decoder}
\end{figure}

To test the generative capacity of the VED model, we isolate the trained decoder and feed in randomly generated codes sampled from the generative model's prior distribution $\mathcal{N}(0, I_r)$ and evaluate the distribution of the decoded output. We see in Figure~\ref{fig:decoder} that the distribution of synthetic outputs more closely resembles the (marginal, feature-wise) distribution of the test data when we train with \emph{both} $\beta$ and $\lambda$ regularization, with the best generalization performance found for $\beta = 0.1, \lambda = 0.1$.
It can also be seen that solely increasing $\beta$ is not sufficient to improve generative performance, which is to be expected as the KL regularization term~\eqref{eq:kld} is similar but not equivalent to penalizing the divergence between the aggregate encoding distribution and the prior distribution. On the other hand, as noted in Section~\ref{sec:ved:disentanglement}, encouraging the aggregate encoding distribution to resemble the prior enhances the generative capacity of the model, which is consistent with Figure~\ref{fig:decoder}.
These results demonstrate that carefully tuning both $\beta$ and $\lambda$ regularization parameters can significantly improve the VED model's generative performance without significantly degrading reconstruction performance. 

\subsubsection{Covariance of Latent Codes}

\begin{figure}[!htb]
    \centering
    \begin{subfigure}[b]{0.24\textwidth}
        \centering
        \includegraphics[width=\textwidth]{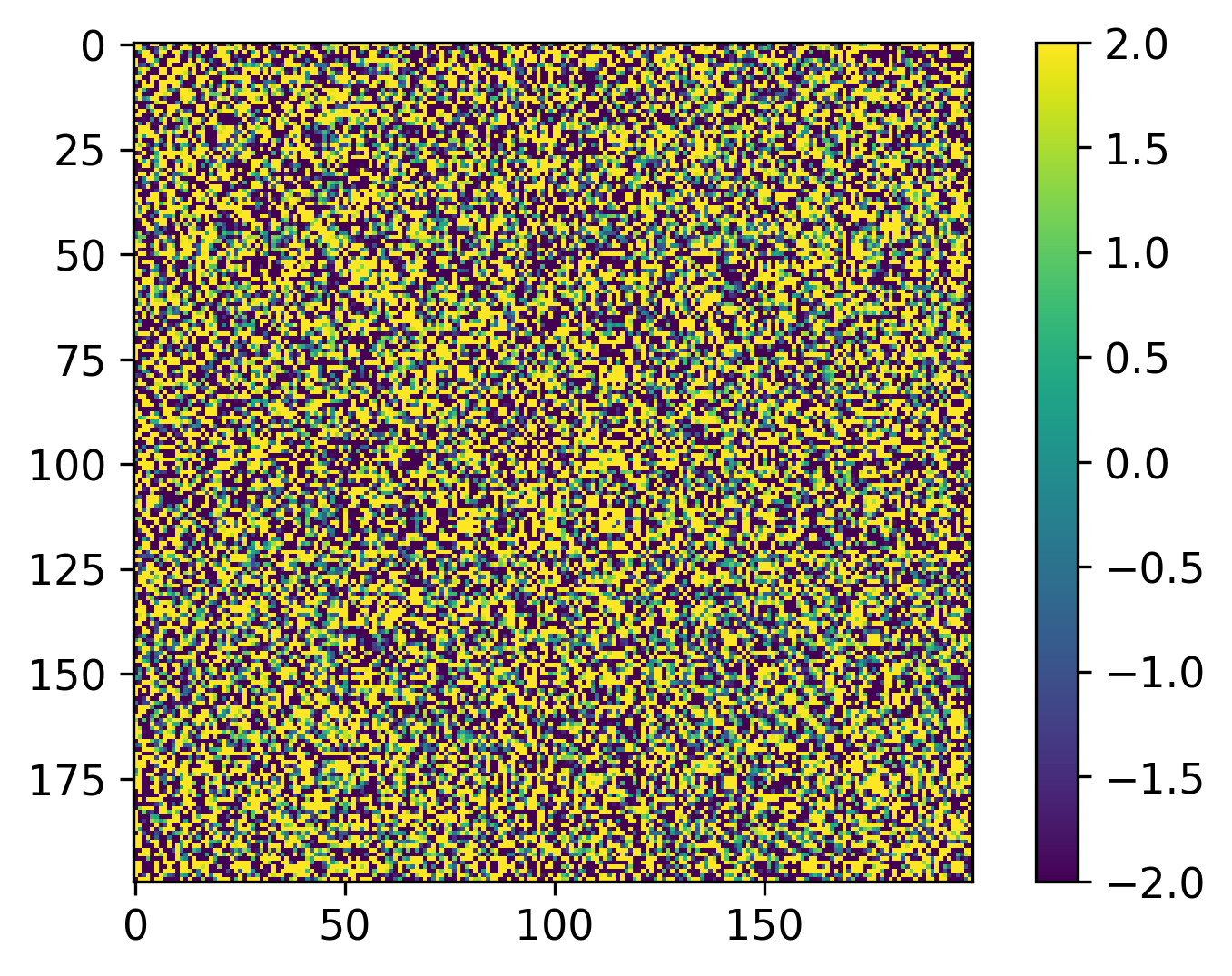}
        \caption{$\beta = 0, \lambda = 0$}
    \end{subfigure}
    \begin{subfigure}[b]{0.24\textwidth}
        \centering
        \includegraphics[width=\textwidth]{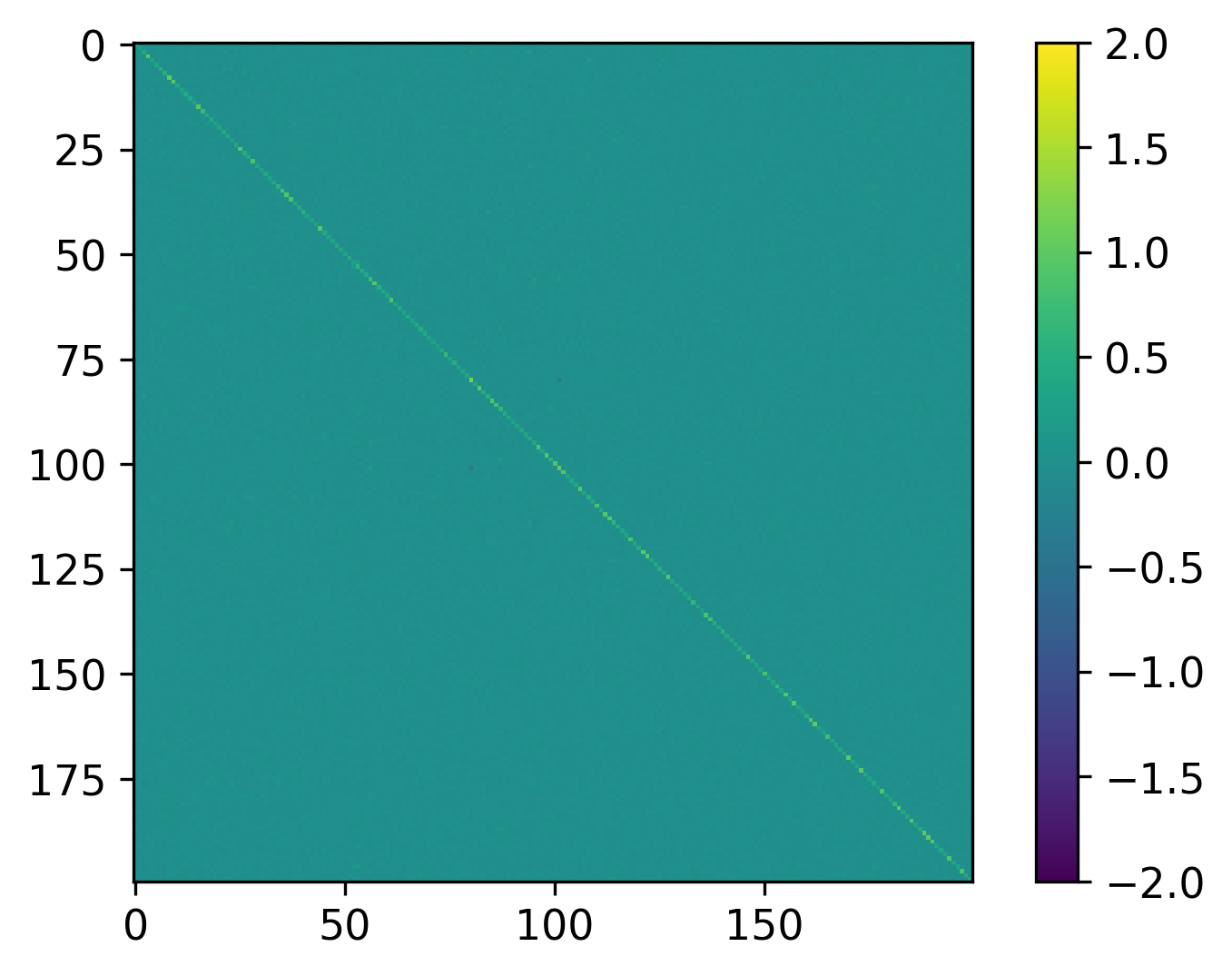}
        \caption{$\beta = 0.01, \lambda = 0.01$}
    \end{subfigure}
    \begin{subfigure}[b]{0.24\textwidth}
        \centering
        \includegraphics[width=\textwidth]{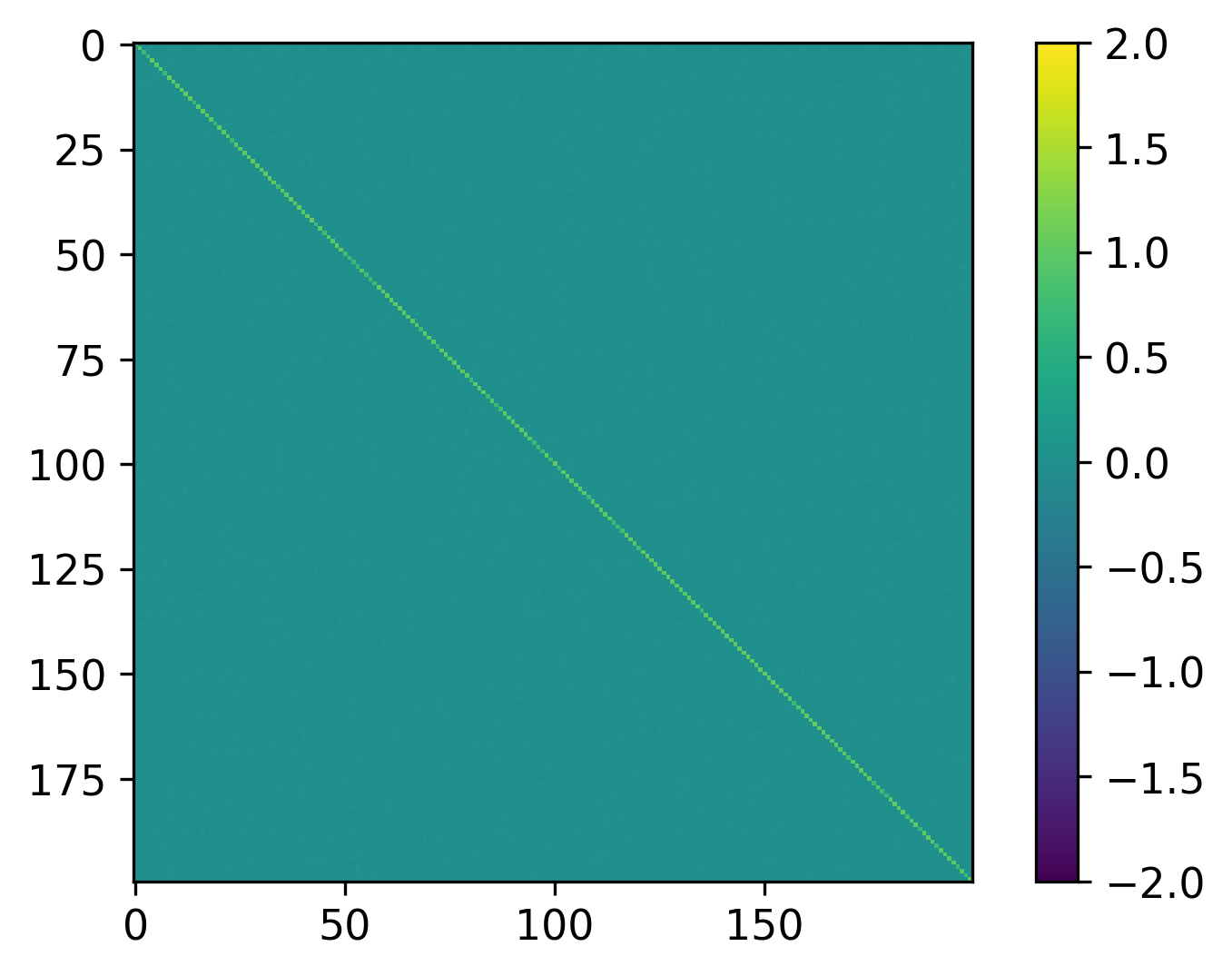}
        \caption{$\beta = 0.1, \lambda = 0.1$}
    \end{subfigure}
    \begin{subfigure}[b]{0.24\textwidth}
        \centering
        \includegraphics[width=\textwidth]{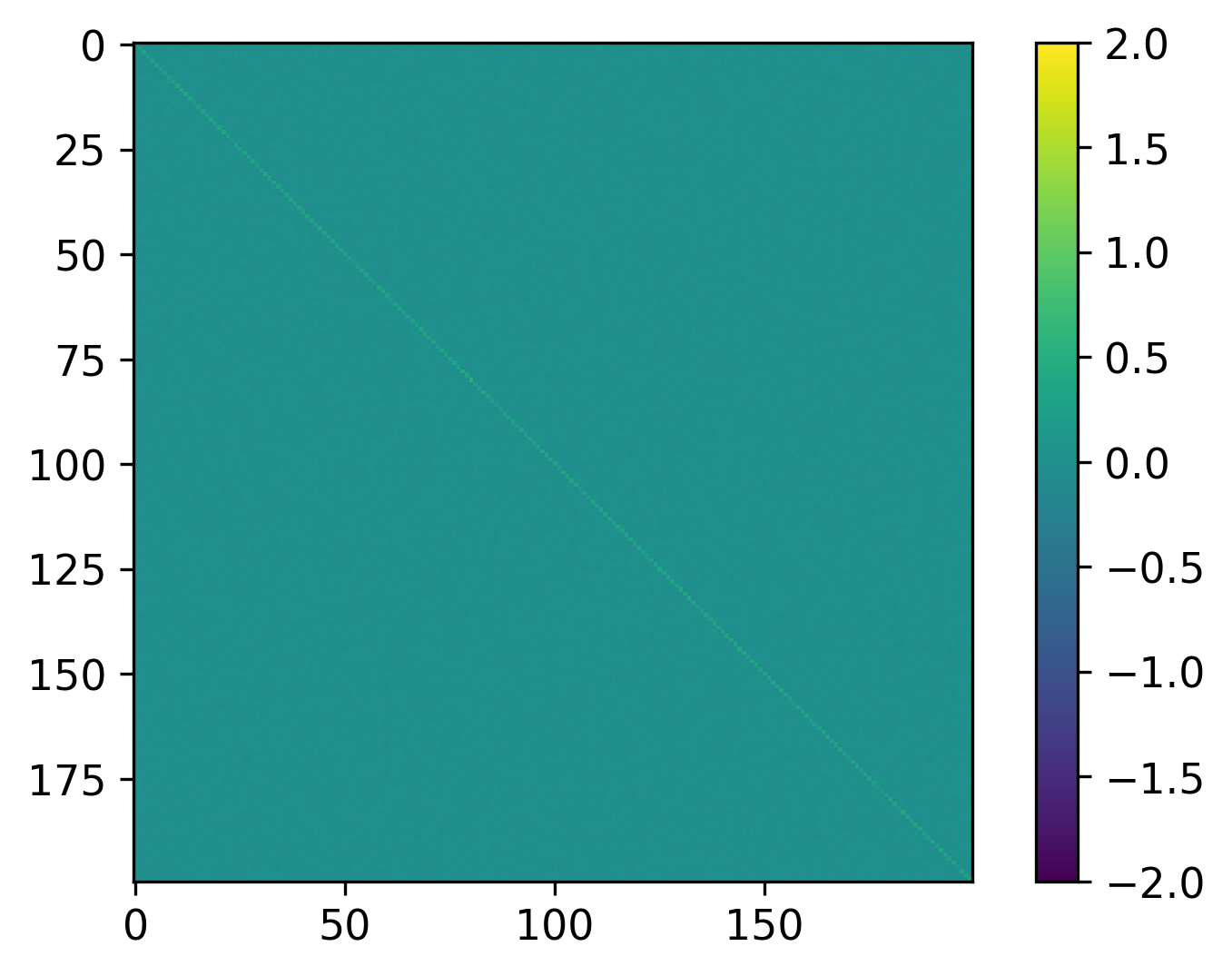}
        \caption{$\beta = 0, \lambda = 0.1$}
    \end{subfigure}
    
    \caption{Empirical covariance of the aggregate encoding distribution computed for $r = 200$ and various choices of $\beta$, $\lambda$.}
    \label{fig:cov1}
\end{figure}

Finally, we analyze the effect of the two regularization terms on the covariance of the aggregate encoding distribution.
For this, we generate an ensemble of codes by sampling from the empirical distribution of test data and employing the ``reparameterization trick'' as
\begin{equation*}
z = g_\theta(x) + \epsilon \odot \exp \tfrac{1}{2} h_\theta(x), \quad \epsilon \sim \mathcal{N}(0, I_r), \quad x \sim \hat{p}(x),
\end{equation*}
then we compute and visualize the empirical covariance of these codes.
For perfectly mutually uncorrelated codes we expect the off-diagonal terms to be zero.
Figure~\ref{fig:cov1} shows the covariances obtained for various choices of $\beta$ and $\lambda$ and $r = 200$. It can be seen that no regularization ($\beta = 0$, $\lambda = 0$) results in a covariance with large off-diagonal terms, which indicates significant correlation between latent features and thus high feature entanglement. On the other hand, the choice $\beta = 0$, $\lambda = 0.1$ results in a covariance that most closely resembles the standard Gaussian prior covariance among all studied choices of $\beta$, $\lambda$, more than just using the covariance penalty $COV(\varphi)$ alone ($\lambda > 0, \beta = 0$). This synergistic effect suggests that the KLD regularization not only shapes the overall latent space distribution but also contributes to the disentanglement of latent features, and that again, the more regularized model is more capable for tasks relating to generating synthetic data from random codes, as it leads to less correlation between codes due to more strongly weighted KLD and covariance regularization.

\section{Discussion and Related Works}
\label{sec:discussion}

Here, we discuss our results in the context of existing literature on VED training, dimensionality reduction, and disentanglement, alongside the most relevant related works in these domains.
Our results demonstrate that the VED framework can effectively capture high-dimensional input-output relationships with a latent dimensionality as low as $r = 50$ while maintaining good reconstruction performance. \citet{https://doi.org/10.1029/2022MS003130} also performed a nonlinear dimensionality reduction task via a VED, however, they used a variation of the $\beta$-VAE loss formulation that was not formally motivated, together with an annealing factor to dynamically weight the KLD-component of the loss based on the number of training epochs in order to improve reconstruction performance. This is similar to \cite{jacobsen2022disentangling} which also investigates the problem of modeling physical datasets using VAEs together with disentanglement-promoting techniques and an annealing $\beta$ schedule over epochs. As our experiment is more motivated by disentanglement (and balancing this metric with MSE), we find no benefit from using annealing schedules on either the KLD and disentanglement-promoting components of our loss, as we are able to achieve low MSE with constant $\beta$ and $\lambda$ values throughout training, and are able to minimize feature entanglement (measured in terms of mutual correlation between latent variables) with our additional covariance regularization term.

In the context of dimensionality reduction, we find that our non-linear VED outperforms the linear CCA's latent dimensionality estimate of $\sim \!\!\!147$. This finding agrees with \citet{serani2018assessing}, which finds that a deep auto-encoder outperforms linear dimensionality reduction approaches like PCA (and nonlinear approaches like kernel PCA). While \cite{serani2018assessing} considers the single-dataset encoding-decoding case with a VAE architecture, our results show this application in the cross-dataset case through the VED framework. Another work that compares dimensionality reduction approaches for modeling fluid dynamics is \citet{Mendez_2023}, however, neither of the nonlinear techniques tested in the paper involve deep learning. We find that our Deep-Learning VED using state-of-the-art architectures (e.g., residual layers, \cite{he2015deepresiduallearningimage}) allows for the best performance for our encoding-decoding task even at minimized latent dimension.

Additionally, our work highlights the synergistic effect of combining KL-divergence ($\beta$) and covariance ($\lambda$) regularization. This combination leads to improved generative capabilities and better-structured latent representations. We motivate this approach by that presented in \citet{mathieu2019disentanglingdisentanglementvariationalautoencoders}, which introduces a composite loss including both $\beta$-weighted KL-regularization and an additional regularization term. After applying this approach with the \cite{kumar2018variational} loss as the secondary regularization term, we saw that the decoder's ability to generate accurate outputs from Gaussian noise improved significantly when multiple types of regularization were applied, and additionally, this training strategy led to more disentangled latent features due to our analysis of the covariance matrices of latent codes. 
Interestingly, our results show that within the constrained ranges of $\beta \in \{ 0, 0.01, 0.1\}$ and $\lambda \in \{ 0, 0.01, 0.1\}$, regularization has either no negative impact or a positive effect on reconstruction performance. This finding contrasts with traditional regularization approaches in machine learning, where it has been noted \cite{tschannen2018recent} that increased regularization often leads to lower reconstruction quality. On the other hand, our results suggest that carefully chosen regularization parameters can improve generative capacity without significantly compromising reconstruction performance. Our decoded Gaussian noise results follow those presented in \citet{jacobsen2022disentangling}, however, while \cite{jacobsen2022disentangling} uses an annealing-$\beta$-VAE loss formulation, we use constant $\beta$ and $\lambda$.

\section{Conclusion}
In this project, we formulated a VED framework for constructing low-dimensional surrogate models of the high-dimensional input-output relationships. We applied the proposed framework to modeling the spatially sparse observable response of a physical system governed by a stationary partial differential equation as a function of the system’s spatially parameter field, with a focus on minimizing the dimensionality of the low-dimensional representation without significant loss of information.

We found that employing a deeper encoder while maintaining a simple, shallow decoder structure, combined with residual blocks on the encoder provided the best reconstruction accuracy while allowing for a minimized low-dimensional representation. Our model's dimensionality reduction was informed by cross-dataset linear encoding experiments using CCA, which allowed us to converge to a testable range of latent dimensionalities for our final VED architectures.
The VED model's performance revealed that we could shrink the latent dimensionality of the representation to as low as $r = 50$ without significantly sacrificing reconstruction accuracy. Additionally, our model retained high performance across different regularization settings, as we find that under constrained ranges for regularization parameters $\beta$ and $\lambda$ for weighting the KL-divergence and covariance respectively, our model does not incur significant loss of reconstruction performance despite added disentanglement regularization, and in some cases, actually benefits in terms of accuracy from the added regularization terms.

Finally, our contributions highlight the importance of regularization in improving generative capabilities. We demonstrated that KL-divergence regularization ($\beta$) and covariance regularization ($\lambda$) synergistically improve both the disentanglement of latent features and the quality of the generated data. We observe this from both the trained decoder's ability to generate accurate outputs from Gaussian noise and the encoder's disentangled/independent feature representations for latent variables (observed through the covariance matrix of latent codes).
This makes our VED model effective for downstream tasks such as uncertainty quantification via generating synthetic data.

\subsection{Future Works}

The encoder architecture employed in this work can be improved by utilizing graph convolutions \cite{kipf2017gcnn} instead of 2-D convolutions. This will obviate the need to map input data to a regular grid using the \texttt{Map2Grid} transformation step.
Another extension is inspired by \citet{kim2024gamma}, which proposes a geometry-informed, curvature-regularized VAE variant to maximize disentanglement of latent features and improve generalization. This approach is found to promote disentanglement and improve robustness, predictive capacity, and generalization capacity of VAEs. This finding is supported by other geometric disentanglement VAE approaches \cite{tatro2020unsupervisedgeometricdisentanglementsurfaces, aumentadoarmstrong2019geometricdisentanglementgenerativelatent}, and thus we anticipate this can also be extended to the VED setting.
Finally, future work will also explore multi-fidelity and multi-modality approaches by leveraging diverse data sources, including experimental data, synthetic datasets, and models with varying resolutions and physics fidelity.

\section{Acknowledgments}
This research was partially supported by the U.S. Department of Energy Office of Science's Advanced Scientific Computing program, and the Office of Workforce Development for Teachers and Scientists under the Science Undergraduate Laboratory Internships Program. Pacific Northwest National Laboratory is operated by Battelle for the DOE under Contract DE-AC05-76RL01830.

\bibliographystyle{plainnat}
\bibliography{bib}

\section{Appendix}

\subsection{Training Results on 10k, 15k Training Dataset Size}
\label{sec:10k-15k}

Table~\ref{tab:10k-table} shows many of the same trends that we observed in the 5k train size experiment presented in Table~\ref{tab:my_label}. We again highlight in bold the best test MSE values with their corresponding KLD per latent dimension $r$. 

\begin{table}[!htb]
    \centering
    \begin{tblr}{
        colspec = {l l | l  l | l  l  | l l | l l |},
  cell{1}{1} = {c = 2,r = 2}{c,m},
  cell{1}{3} = {c = 2}{halign = c},
  cell{1}{5} = {c = 2}{halign = c},
  cell{1}{7} = {c = 2}{halign = c},
  cell{3}{1} = {r = 3}{valign = m},
  cell{6}{1} = {r = 3}{valign = m},
  cell{9}{1} = {r = 3}{valign = m},
  cell{12}{1} = {r = 3}{valign = m},
  hline{3}   = {black},
  hline{4}   = {black, dotted},
  hline{5}   = {black, dotted},
  vline{4}   = {black, dotted},
  hline{6}   = {black},
  hline{7}   = {black, dotted},
  hline{8}   = {black, dotted},
  vline{6}   = {black, dotted},
  hline{9}   = {black},
  hline{10}   = {black, dotted},
  hline{11}   = {black, dotted},
  vline{8}   = {black, dotted},
  hline{12}   = {black},
  hline{13}   = {black, dotted},
  hline{14}   = {black, dotted},
  hline{15}   = {black}
  }
  &  &  $\lambda = 0$ & & $\lambda = 0.01$ & & $\lambda = 0.1$ & \\
  &  &  MSE & KLD & MSE & KLD & MSE & KLD\\
  $r = 50$ &  $\beta = 0$ &  0.0313 & 55.852 & 0.0275 & 7.5223 & \textbf{0.0274} & \textbf{7.3724}\\
  &  $\beta = 0.01$ &  0.0293 & 0.7113 & 0.0280 & 2.8037 & 0.0280 & 2.8745\\
  &  $\beta = 0.1$ &  0.0341 & 1.4816 & 0.0338 & 1.4716 & 0.0347 & 1.4326\\
  $r = 100$ &  $\beta = 0$ &  0.0217 & 43.876 & \textbf{0.0170} & \textbf{7.9208} & 0.0179 & 8.4038\\
  &  $\beta = 0.01$ &  0.0196 & 2.2494 & 0.0189 & 2.3165 & 0.0198 & 2.2615\\
  &  $\beta = 0.1$ &  0.0310 & 0.7882 & 0.0312 & 0.7698 & 0.0336 & 0.7207\\
  $r = 150$ &  $\beta = 0$ &  0.0212 & 33.791 & \textbf{0.0163} & \textbf{7.8363} & 0.0172 & 13.555\\
  &  $\beta = 0.01$ &  0.0205 & 1.7950 & 0.0186 & 1.7927 & 0.0195 & 1.5030\\
  &  $\beta = 0.1$ &  0.0310 & 0.5153 & 0.0320 & 0.4965 & 0.0347 & 0.4664\\
  $r = 200$ &  $\beta = 0$ & 0.0201 & 34.934 & \textbf{0.0160} & \textbf{7.8401} & 0.0173 & 13.295\\
  &  $\beta = 0.01$ & 0.0204 & 1.3774 & 0.0186 & 1.3271 & 0.0196 & 1.1137\\
  &  $\beta = 0.1$ & 0.0308 & 0.3834 & 0.0318 & 0.3722 & 0.0361 & 0.3390\\
    \end{tblr}
    \caption{Reconstruction and KLD test loss values for $\beta \in \{0, 0.01, 0.1\}, \lambda \in \{0, 0.01, 0.1\}, r \in \{50, 100, 150, 200\}$ on training dataset with 10,000 samples.}
    \label{tab:10k-table}
\end{table}

Specifically, we note that there is improved reconstruction accuracy from higher latent dimensions, but $r = 50$ still yields very low test MSE values. We can see as before that the test KLD decreases with increasing latent dimension and increasing $\beta$. For the most part, we can also see that the test KLD decreases with increasing $\lambda$. In all but one case of possible $\beta$ values, training with $\lambda \neq 0$ decreases the KLD as compared to the same $\beta$ value with $\lambda = 0$. As opposed to our finding from the 5k train size, we find the best MSE performance across all values of $r$ with $\beta = 0$. As before, we use the same range of $\beta$ values for testing ($\beta \in \{0, 0.01, 0.1\}$), as we start to see an appreciable loss in reconstruction accuracy at $\beta = 0.1$. With respect to the $\lambda$ parameter, we see that like before the best test MSE occurs with nonzero $\lambda$, mostly $\lambda = 0.01$, with $\beta = 0$ across all $r$ values. Additionally, Table \ref{tab:10k-table} shows that for large $\beta$ ($\beta = 0.1$), adding the $\lambda$ parameter causes over-regularization and starts to decrease the MSE. As before we see the best synergy between $\lambda$ and $\beta$ while selecting $\lambda = 0.01$ and $\beta = 0.01$  to give us a good balance between the MSE and KLD test metrics.

\begin{table}[!htb]
    \centering
    \begin{tblr}{
        colspec = {l l | l  l | l  l  | l l |},
        cell{1}{1} = {c = 2, r = 2}{c,m},
        cell{1}{3} = {c = 2}{halign = c},
        cell{1}{5} = {c = 2}{halign = c},
        cell{1}{7} = {c = 2}{halign = c},
        cell{3}{1} = {r = 3}{valign = m},
        cell{6}{1} = {r = 3}{valign = m},
        cell{9}{1} = {r = 3}{valign = m},
        cell{12}{1} = {r = 3}{valign = m},
        hline{3} = {black},
        hline{4} = {black, dotted},
        hline{5} = {black, dotted},
        vline{4} = {black, dotted},
        hline{6} = {black},
        hline{7} = {black, dotted},
        hline{8} = {black, dotted},
        vline{6} = {black, dotted},
        hline{9} = {black},
        hline{10} = {black, dotted},
        hline{11} = {black, dotted},
        vline{8} = {black, dotted},
        hline{12} = {black},
        hline{13} = {black, dotted},
        hline{14} = {black, dotted},
        hline{15} = {black}
    }
    &  &  $\lambda = 0$ & & $\lambda = 0.01$ & & $\lambda = 0.1$ & \\
    &  &  MSE & KLD & MSE & KLD & MSE & KLD \\
    $r = 50$ &  $\beta = 0$ &  0.0254 & 51.355 & \textbf{0.0236} & \textbf{7.6194} & 0.0238 & 7.7173 \\
    &  $\beta = 0.01$ &  0.0245 & 2.7898 & 0.0245 & 2.8212 & 0.0244 & 2.8764 \\
    &  $\beta = 0.1$ &  0.0309 & 1.4868 & 0.0310 & 1.4745 & 0.0317 & 1.4274 \\
    $r = 100$ &  $\beta = 0$ &  0.0158 & 50.363 & \textbf{0.0137} & \textbf{7.4504} & 0.0145 & 7.8941 \\
    &  $\beta = 0.01$ &  0.0156 & 2.2455 & 0.0153 & 2.3166 & 0.0165 & 2.2724 \\
    &  $\beta = 0.1$ &  0.0280 & 0.7845 & 0.0290 & 0.7663 & 0.0315 & 0.7153 \\
    $r = 150$ &  $\beta = 0$ &  0.0148 & 45.381 & \textbf{0.0123} & \textbf{7.5889} & 0.0135 & 14.313 \\
    &  $\beta = 0.01$ &  0.0156 & 1.7873 & 0.0152 & 1.7974 & 0.0165 & 1.5017 \\
    &  $\beta = 0.1$ &  0.0281 & 0.5186 & 0.0295 & 0.4993 & 0.0330 & 0.4613 \\
    $r = 200$ &  $\beta = 0$ &  0.0145 & 41.893 & \textbf{0.0123} & \textbf{7.8024} & 0.0132 & 14.739 \\
    &  $\beta = 0.01$ &  0.0158 & 1.3707 & 0.0154 & 1.3248 & 0.0163 & 1.1216 \\
    &  $\beta = 0.1$ &  0.0287 & 0.3809 & 0.0297 & 0.3725 & 0.0343 & 0.3386 \\
    \end{tblr}
    \caption{Reconstruction and KLD test loss values for $\beta \in \{0, 0.01, 0.1\}, \lambda \in \{0, 0.01, 0.1\}, r \in \{50, 100, 150, 200\}$ on training dataset with 15,000 samples.}
    \label{tab:15k-table}
\end{table}

We also report our findings for the same tests on the 15k dataset size in Table~\ref{tab:15k-table}. 
Here we again observe many of the same findings: $r = 50$ still yields low test MSE values comparable to $r = 200$ even at this larger dataset size. Additionally, as before, test KLD decreases as we increase $r$, and observed test KLD decreases for larger $\beta$. The effects of $\lambda$ weighted covariance regularization are mostly the same, where test KLD decreases with nonzero $\lambda$, in all but two cases. Also, nonzero $\lambda$ (specifically $\lambda = 0.01$) and $\beta = 0$ consistently leads to the lowest test MSE across all four $r$ values. Unlike before, we can see that for $\beta = 0.1$, adding the $\lambda$ parameter does decrease the MSE, so the added $\lambda$ does not over-regularize on the 15k dataset. Finally, we maintain that the best combination of weights for $\lambda$ and $\beta$ is $\lambda = 0.01$ and $\beta = 0.01$ to best balance the MSE and KLD test metrics.

\end{document}